\documentclass{article}

\PassOptionsToPackage{numbers, compress}{natbib}

\usepackage[final]{neurips_2022}

\usepackage[utf8]{inputenc} 
\usepackage[T1]{fontenc}    
\usepackage{hyperref}       
\usepackage{url}            
\usepackage{booktabs}       
\usepackage{amsfonts}       
\usepackage{nicefrac}       
\usepackage{microtype}      
\usepackage{xcolor}         
\usepackage{graphicx}

\usepackage{amsmath}
\usepackage{amssymb}

\usepackage{algpseudocode}
\usepackage{algorithm}

\algnewcommand\algorithmicforeach{\textbf{for each:}}
\algnewcommand\ForEach{\item[ \algorithmicforeach]}

\usepackage{cleveref}

\usepackage{placeins}
\usepackage{stfloats}
\usepackage{multicol}

\usepackage{array}
\usepackage{mathtools}
\usepackage{relsize}

\newcommand{\vectorproj}[2][]{\textit{proj}_{#1}#2}

\newcommand{\ccbyncsa}{\href{https://creativecommons.org/licenses/by-nc-sa/4.0/}{CC BY-NC-SA 4.0}}

\newcommand{\nvsrc}{\href{https://nvlabs.github.io/stylegan2/license.html
}{Nvidia Source Code License-NC}}

\newcommand{\mitlic}{\href{https://opensource.org/licenses/MIT}{MIT License}}

\newcommand{\adblic}{\href{https://github.com/utkarshojha/few-shot-gan-adaptation/blob/main/LICENSE.txt}{Adobe Research License}}

\title{HyperDomainNet: Universal Domain Adaptation for Generative Adversarial Networks}

\author{%
  Aibek Alanov\thanks{Equal contribution}\ \ \thanks{Corresponding author} \\
  HSE University, AIRI\thanks{Artificial Intelligence Research Institute} \\
  Moscow, Russia \\
  \texttt{alanov.aibek@gmail.com} \\
  \And 
  Vadim Titov\footnotemark[1]\ \ \footnotemark[2] \\
  MIPT\thanks{Moscow Institute of Physics and Technology}, 
  AIRI\footnotemark[3] \\
  Moscow, Russia \\
  \texttt{titov.vn@phystech.edu} \\
  \And 
  Dmitry Vetrov \\
  HSE University, AIRI\footnotemark[3] \\
  Moscow, Russia \\
  \texttt{vetrovd@yandex.ru} \\
}

\begin{document}

\maketitle

\begin{abstract}
Domain adaptation framework of GANs has achieved great progress in recent years as a main successful approach of training contemporary GANs in the case of very limited training data. In this work, we significantly improve this framework by proposing an extremely compact parameter space for fine-tuning the generator. We introduce a novel domain-modulation technique that allows to optimize only 6 thousand-dimensional vector instead of 30 million weights of StyleGAN2 to adapt to a target domain. We apply this parameterization to the state-of-art domain adaptation methods and show that it has almost the same expressiveness as the full parameter space. Additionally, we propose a new regularization loss that considerably enhances the diversity of the fine-tuned generator. Inspired by the reduction in the size of the optimizing parameter space we consider the problem of multi-domain adaptation of GANs, i.e. setting when the same model can adapt to several domains depending on the input query. We propose the HyperDomainNet that is a hypernetwork that predicts our parameterization given the target domain. We empirically confirm that it can successfully learn a number of domains at once and may even generalize to unseen domains. Source code can be found at \href{https://github.com/MACderRu/HyperDomainNet}{this github repository}.
\end{abstract}

\section{Introduction}
Contemporary generative adversarial networks (GANs) \cite{goodfellow2014generative, karras2019style, karras2020analyzing, karras2020training, brock2018large} show remarkable performance in modeling image distributions and have applications in a wide range of computer vision tasks (image enhancement \cite{ledig2017photo, yang2021gan}, editing \cite{harkonen2020ganspace, shen2020interpreting}, image-to-image translation \cite{isola2017image, zhu2017unpaired, zhu2017toward}, etc.). However, the training of modern GANs requires thousands of samples that limits its applicability only to domains that are represented by a large set of images. The mainstream approach to sidestep this limitation is transfer learning (TL), i.e. fine-tuning the generative model to a domain with few samples starting with a pretrained source model. 

The standard approach of GAN TL methods is to fine-tune almost \emph{all} weights of the pretrained model \cite{li2020few, mo2020freeze, wang2018transferring, wang2020minegan, karras2020training, zhao2020differentiable, ojha2021few, gal2021stylegan, zhu2021mind}. It can be reasonable in the case when the target domain is very far from the source one, e.g. when we adapt the generator pretrained on human faces to the domain of animals or buildings. However, there is a wide range of cases when the distance between data domains is not so far. In particular, the majority of target domains used in works \cite{li2020few, wang2020minegan, ojha2021few, gal2021stylegan, zhu2021mind} are similar to the source one and differ mainly in texture, style, geometry while keep the same content like faces or outdoor scenes. 
For such cases it seems redundant to fine-tune all weights of the source generator. It was shown in the paper \cite{wu2021stylealign} that after transfer learning of the StyleGAN2 \cite{karras2020analyzing} to similar domains some parts of the network almost do not change. This observation motivates us to find a more efficient and compact parameter space for domain adaptation of GANs. 

In this paper, we propose a novel \emph{domain-modulation} operation that reduces the parameter space for fine-tuning the StyleGAN2. The idea is to optimize for each target domain only a single vector $d$. We incorporate this vector into the StyleGAN2 architecture through the modulation operation at each convolution layer. The dimension of the vector $d$ equals 6 thousand that is \emph{5 thousand times} less than the original weights space of the StyleGAN2. We apply this parameterization for the state-of-the-art domain adaptation methods StyleGAN-NADA \cite{gal2021stylegan} and MindTheGAP \cite{zhu2021mind}. We show that it has almost the same expressiveness as the full parameterization while being more lightweight. To further advance the domain adaptation framework of GANs we propose a new regularization loss that improves the diversity of the fine-tuned generator.

Such considerable reduction in the size of the proposed parameterization motivates us to consider the problem of multi-domain adaptation of GANs, i.e. when the same model can adapt to multiple domains depending on the input query. Typically, this problem is tackled by previous methods just by fine-tuning separate generators for each target domain independently. In contrast, we propose to train a hyper-network that predicts the vector $d$ for the StyleGAN2 depending on the target domain. We call this network as HyperDomainNet. Such hyper-network would be impossible to train if we needed to predict all weights of StyleGAN2. 
The immediate benefits of multi-domain framework consist of reducing the training time and the number of trainable parameters because instead of fine-tuning $n$ separate generators we train one HyperDomainNet to adapt to $n$ domains simultaneously. Another advantage of this method is that it can generalize to \emph{unseen} domains if $n$ is sufficiently large and we empirically observe this effect. 

We provide extensive experiments to empirically confirm the effectiveness of the proposed parameterization and the regularization loss on a wide range of domains. We illustrate that our parameterization can achieve quality comparable with the full parameterization (i.e. when we optimize all weights). The proposed regularization loss significantly improves the diversity of the fine-tuned generator that is validated qualitatively and quantitatively. Further, we conduct experiments with the HyperDomainNet and show that it can be successfully trained on a number of target domains simultaneously. Also we show that it can generalize to a number of diverse unseen domains.

To sum up, our main contributions are
\begin{itemize}
    \item We reduce the number of trainable parameters for domain adaptation of StyleGAN2 \cite{karras2020analyzing} generator by proposing the domain-modulation technique. Instead of fine-tuning all 30 millions weights of StyleGAN2 for each new domain now we can train only 6 thousand-dimensional vector.
    \item We introduce a novel regularization loss that considerably improves the diversity of the adapted generator. 
    \item We propose a HyperDomainNet that predicts the parameterization vector for the input domain and allows multi-domain adaptation of GANs. It shows inspiring generalization results on unseen domains. 
\end{itemize}


\section{Related Work}

\paragraph{Domain Adaptation of GANs} 
The aim of few-shot domain adaptation of GANs is to learn accurate and diverse distribution of the data represented by only a few images. The standard approach is to utilize a generator pretrained on source domain and fine-tune it to a target domain. There are generally two different regimes of this task. The first one is when we adapt the generator to completely new data (e.g. faces $\rightarrow$ landscapes, churches, etc.), and the second regime is when the target domain relates to the source one (e.g. faces $\rightarrow$ sketches, artistic portraits, etc.). 

Methods that tackle the first regime typically require several hundreds or thousands samples to adapt successfully. Such setting assumes that the weights of the generator should be changed significantly because the target domain can be very far from the source. The paper \cite{karras2020training} shows that for distant domains training from scratch gives comparable results to transfer learning. It also confirms that for such regime there is no point to reduce the parameter space. Typcially such approaches utilize data augmentations \cite{karras2020training, tran2021data, zhao2020differentiable, zhao2020image}, or use auxiliary tasks for the discriminator to more accurately fit the available data \cite{liu2020towards, yang2021data}, or freeze lower layers of the discriminator to avoid overfitting \cite{mo2020freeze}. Another standard techniques for the effective training of GANs is to apply different normalization methods \cite{miyato2018spectral, kurach2019large, senderovich2022towards} to stabilize the training process. 

In the second regime the transfer learning is especially crucial because the pretrained generator already contains many information about the target domain. In this setting the required number of available data can be significantly smaller and range from hundreds to several images. The main challenges in the case of such limited data are to avoid over-fitting of the generator and leverage its diversity learned from the source domain. To tackle these challenges existing methods introduce restrictions on the parameter space \cite{robb2020few, noguchi2019image}, mix the weights of the adapted and the source generators \cite{pinkney2020resolution}, utilize a small network to force sampling in special regions of the latent space \cite{wang2020minegan}, propose new regularization terms \cite{li2020few, tseng2021regularizing}, or apply contrastive learning techniques to enhance cross-domain consistency \cite{ojha2021few}. The state-of-the-art methods \cite{gal2021stylegan, zhu2021mind} leverage supervision from vision-language CLIP model \cite{radford2021learning}. StyleGAN-NADA \cite{gal2021stylegan} applies it for text-based domain adaptation when we have no access to images but only to the textual description. MindTheGap \cite{zhu2021mind} employs CLIP model to further significantly improve the quality of one-shot domain adaptation.

\paragraph{Constraining Parameter Space for GAN's Adaptation}
In the second regime of GAN's adaptation it is especially important for the generator to leverage the information from the source domain during adapting to the target one. The common approach is to introduce some restrictions on the trainable weights to regularize them during fine-tuning. For example, the work \cite{robb2020few} proposes to optimize only the singular values of the pretrained weights and apply it for few shot domain adaptation, however the reported results show the limited expressiveness of such parameterization \cite{robb2020few}. Another method \cite{noguchi2019image} constrains the parameter space for models with batch normalization (BN) layers such as BigGAN \cite{brock2018large} by optimizing only BN statistics during fine-tuning. While it allows to decrease the number of trainable parameters, it also considerably reduces the expressiveness of the generator \cite{robb2020few, ojha2021few}. Other approach is to adaptively choose a subset of layers during optimization at each step as in StyleGAN-NADA \cite{gal2021stylegan}. It helps to stabilize the training, however it does not reduce the parameter space because each layer can potentially be fine-tuned. 
The alternative method is to optimize parameters in the latent space of StyleGAN as in TargetCLIP \cite{chefer2021image} and the size of this space is much smaller than the size of the whole network. However, such approach works mainly for in-domain editing and shows poor quality in adapting to new domains.
In contrast, our parameterization has the comparable expressiveness and adaptation quality as the full parameter space while its size is less by three orders of magnitude. 


\section{Preliminaries}\label{sec:stylegan}
In this work, we focus on StyleGAN generators in the context of domain adaptation. We consider StyleGAN2 \cite{karras2020analyzing} as a base model. As the state-of-the-art domain adaptation methods we use StyleGAN-NADA \cite{gal2021stylegan} and MindTheGAP \cite{zhu2021mind}.

\paragraph{StyleGAN2}
The StyleGAN2 generation process consists of several components. The first part is a mapping network $M(z)$ that takes as an input random vectors $z \in \mathcal{Z}$ from the initial latent space, $\mathcal{Z}$ that is typically normally distributed. It transforms these vectors $z$ into the intermediate latent space $\mathcal{W}$. 
Each vector $w \in \mathcal{W}$ is further fed into different affine transformations $A(w)$ for each layer of the generator. The output of this part forms StyleSpace $\mathcal{S}$ \cite{wu2021stylespace} that consists of channel-wise style parameters $s = A(w)$. The next part of the generation process is the synthesis network $G_{sys}$ that takes as an input the constant tensor $c$ and style parameters $s$ at the corresponding layers and produces the final feature maps at different resolutions $F = G_{sys}(c, s)$. These feature maps move on to the last part which consists of toRGB layers $G_{tRGB}$ that generate the output image $I = G_{tRGB}(F)$.

\paragraph{Problem Formulation of Domain Adaptation}
The problem of domain adaptation of StyleGAN2 can be formulated as follows. We are given a trained generator $G^A$ for the source domain $A$, and the target domain $B$ that is represented by the one image $I_B$ (one-shot adaptation) or by the text description $t_B$ (text-guided adaptation). The aim is to fine-tune the weights $\theta$ of a new generator $G^B_{\theta}$ for the domain $B$ starting from the weights of $G^A$. The optimization process in the general form is
\begin{gather}
    \mathcal{L}_B(\theta) = \mathcal{L}(\{G^B_{\theta}(w_i)\}_{i=1}^n, \{G^A(w_i)\}_{i=1}^n, G^B_{\theta}, B, A) \; \rightarrow \; \min_{\theta}, \label{eq:general}
\end{gather}
where $\mathcal{L}$ is some loss function, $n$ is a batch size, $w_1, \dots, w_n$ are random latent codes, $\{G^B_{\theta}(w_i)\}_{i=1}^n$ and $\{G^A(w_i)\}_{i=1}^n$ are batches of images sampled by $G^B_{\theta}$ and $G^A$ generators, respectively, and $B, A$ are domains that are represented by images or text descriptions. 

\paragraph{CLIP model} CLIP \cite{radford2021learning} is a vision-language model that is composed of text and image encoders $E_T$, $E_I$, respectively, that maps their inputs into a joint, multi-modal space of vectors with a unit norm (this space is often called as CLIP space). In this space the cosine distance between embeddings reflects the semantic similarity of the corresponding objects. 

\paragraph{StyleGAN-NADA}
StyleGAN-NADA \cite{gal2021stylegan} is a pioneering work that utilizes the CLIP model \cite{radford2021learning} for text-guided domain adaptation of StyleGAN. 
The proposed loss function is
\begin{gather}
    \Delta T (B, A) = E_T(t_{B}) - E_T(t_{A}), \nonumber \\ 
    \Delta I(G^{B}_{\theta}(w), G^{A}(w)) = E_I(G^{B}_{\theta}(w)) - E_I(G^{A}(w)), \nonumber
     \\
    \mathcal{L}_{direction}(G^{B}_{\theta}(w), G^{A}(w), B, A) = 1 - \dfrac{\Delta I(G^{B}_{\theta}(w), G^{A}(w)) \cdot \Delta T(B, A)}{|\Delta I(G^{B}_{\theta}(w), G^{A}(w))| |\Delta T(B, A)|}.
\label{eq:direction}
\end{gather}
The idea is to align the CLIP-space direction between the source and target images $\Delta I(G^{B}_{\theta}(w), G^{A}(w))$ with the direction between a pair of source and target text descriptions $\Delta T (B, A)$. 
So, the overall optimization process has the form
\begin{gather}
    \mathcal{L}_B(\theta) = \sum\limits_{i=1}^n \mathcal{L}_{direction}(G^{B}_{\theta}(w_i), G^{A}(w_i), B, A) \; \rightarrow \; \min_{\theta}.
\end{gather}
In StyleGAN-NADA method the $\mathcal{L}_B(\theta)$ loss is optimized only with respect to the weights $\theta$ of the synthesis network $G^{B}_{sys}$ which has 24 million weights. 

\paragraph{MindTheGap}
The MindTheGap method \cite{zhu2021mind} is proposed for a one-shot domain adaptation of StyleGAN, i.e. the domain $B$ is represented by the single image $I_B$. 
In principle StyleGAN-NADA method can solve this problem just by replacing the text direction $\Delta T(B, A)$ from \Cref{eq:direction} to an image one 
\begin{gather}
    \Delta I'(B, A) = E_I(I_B) - \dfrac{1}{|A|}\sum\limits_{I_A \in A}[E_I(I_A)], \label{eq:im_dir}
\end{gather}
where $\dfrac{1}{|A|}\sum\limits_{I_A \in A}[E_I(I_A)]$ is the mean embedding of the images from domain $A$. However, as stated in \cite{zhu2021mind} this leads to an undesirable effect that transferred images lose the initial diversity of domain $A$ and become too close to the $I_B$ image. So, the key idea of the MindTheGap is to replace the mean embedding from \Cref{eq:im_dir} by the embedding of projection $I_A^*$ of $I_B$ image to $A$ domain obtained by the GAN inversion method II2S \cite{zhu2020improved}:
\begin{gather}
    \Delta I''(B, A) = E_I(I_B) - E_I(I_A^*),
\end{gather}
So, the MindTheGap uses the modified $\mathcal{L}'_{direction}$ loss that is renamed to $\mathcal{L}_{clip\_accross}$
\begin{gather}
    \mathcal{L}_{clip\_accross}(G^{B}_{\theta}(w), G^{A}(w), B, A) = 1 - \dfrac{\Delta I(G^{B}_{\theta}(w), G^{A}(w)) \cdot \Delta I''(B, A)}{|\Delta I(G^{B}_{\theta}(w), G^{A}(w))| |\Delta I''(B, A)|}. \label{eq:clip_across}
\end{gather}
In addition to this idea several new regularizers are introduced that force the generator $G_{\theta}^B$ to reconstruct the $I_B$ image from its projection $I_A^*$. It further stabilizes and improves the quality of domain adaption. Overall, the MindTheGAP loss function $\mathcal{L}_{MTG}$ has four terms to optimize $G_{\theta}^B$.
For more details about each loss please refer to the original paper \cite{zhu2021mind}. 

\section{Approach}

\subsection{Domain-Modulation Technique for Domain Adaptation} \label{sec:domain-modulation}
Our primary goal is to improve the domain adaptation of StyleGAN by exploring an effective and compact parameter space to use it for fine-tuning $G^B_{\theta}$. 
As we described in \Cref{sec:stylegan} StyleGAN has four components: the mapping network $M(\cdot)$, affine transformations $A(\cdot)$, the synthesis network $G_{sys}(\cdot, \cdot)$, and toRGB layers $G_{tRGB}(\cdot)$. It is observed in the paper \cite{wu2021stylealign} that the main part of StyleGAN that is mostly changed during fine-tuning to a target domain is the synthesis network $G_{sys}(\cdot, \cdot)$. It is also confirmed by StyleGAN-NADA \cite{gal2021stylegan} and MindTheGap \cite{zhu2021mind} methods as they adapt only the weights of $G_{sys}(\cdot, \cdot)$ for the target domain.

So, we aim to find an effective way to fine-tune the weights of feature convolutions of $G_{sys}(\cdot, \cdot)$. In StyleGAN2 \cite{karras2020analyzing} these convolutions utilize modulation/demodulation operations to process the input tensor and the corresponding style parameters $s$. Let us revisit the mechanism of these operations:
\begin{align}
    \text{modulation: } w'_{ijk} &= s_i\cdot w_{ijk}, \label{eq:modulation} \\
    \text{demodulation: } w''_{ijk} &= \dfrac{w'_{ijk}}{\sqrt{\sum\limits_{i,k}{w'_{ijk}}^2 + \varepsilon}}, 
\end{align}
where $w, w'$ and $w''$ are the original, modulated and demodulated weights, respectively, $s_i$ is the component of the style parameters $s$, $i$ and $j$ enumerate input and output channels, respectively.
The idea behind modulation/demodulation is to replace the standard adaptive instance normalization (AdaIN) \cite{ulyanov2016instance, dumoulin2016learned} to a normalization that is based on the expected statistics of the input feature maps rather than forcing them explicitly \cite{karras2020analyzing}. So, the modulation part is basically an adaptive scaling operation as in AdaIN that is controlled by the style parameters $s$. This observation inspires us to use this technique for the domain adaptation. 

The problem of fine-tuning GANs to a new domain is very related to the task of style transfer where the goal is also to translate images from the source domain to a new domain with the specified style. The contemporary approach to solve this task is to train an image-to-image network which takes the target style as an input condition. The essential ingredient of such methods is the AdaIN that provides an efficient conditioning mechanism. In particular, it allows to train arbitrary style transfer models \cite{huang2017arbitrary}. So, it motivates us to apply the AdaIN technique for adapting GANs to new domains. 

\begin{figure}
  \includegraphics[width=\linewidth]{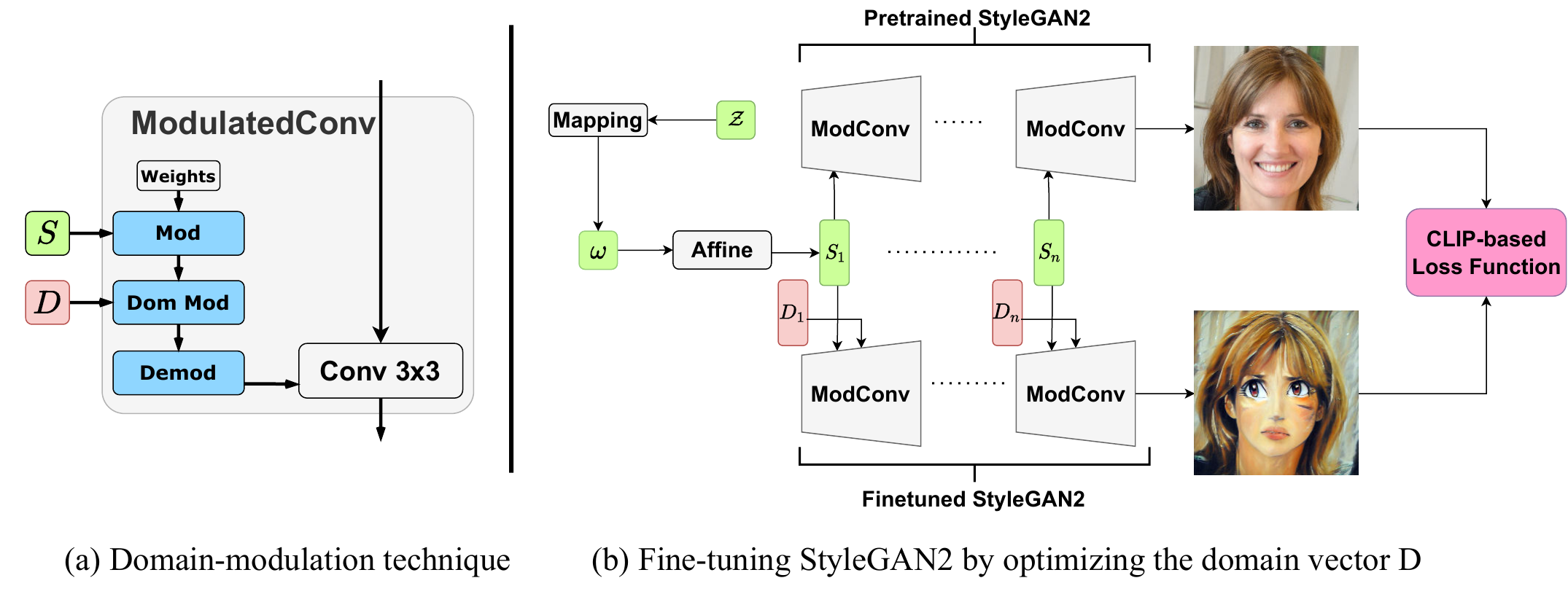}
  \caption{Detailed diagram of proposed method. (a) Revised ModulatedConv block with introduced domain-modulation operation. (b) Fully detailed training process of the domain adaptation with the proposed domain-modulation technique.}
  \label{fig:training_diagram}
\end{figure}

We introduce a new \emph{domain-modulation} operation that reduces the parameter space for fine-tuning StyleGAN2. The idea is to optimize only a vector $d$ with the same dimension as the style parameters $s$. We incorporate this vector into StyleGAN architecture by the additional modulation operation after the standard one from \Cref{eq:modulation}:
\begin{align}
    \text{domain-modulation: } w'_{ijk} &= d_i\cdot w_{ijk}, \label{eq:domain-mod} 
\end{align}
where $d_i$ is the component of the introduced domain parameters $d$ (see \Cref{fig:training_diagram}a). So, instead of optimizing all weights $\theta$ of the $G_{sys}$ part we train only the vector $d$. 

We apply these new parameterization to StyleGAN-NADA and MindTheGAP methods, i.e. instead of optimizing its loss functions wrt $\theta$ we optimize it wrt $d$ vector (see \Cref{fig:training_diagram}b)
The dimension of the vector $d$ equals 6 thousand that is 4 thousand times less than the original weights space $\theta$ of $G_{sys}(\cdot, \cdot)$ part. While the proposed parameter space is radically more constrained we observe that it has the expressiveness comparable with the whole weight space. 

\subsection{Improving Diversity of CLIP-Guided Domain Adaptation}
The CLIP-based domain adaptation methods StyleGAN-NADA and MindTheGap use $\mathcal{L}_{direction}$ (or $\mathcal{L}_{clip\_accross}$) loss (see \Cref{eq:direction,eq:clip_across}) that was initially introduced to deal with the mode collapsing problem of the fine-tuned generator \cite{gal2021stylegan}. However, we empirically observe that it solves the issue only partially. In particular, it preserves the diversity only at the beginning of the fine-tuning process and starts collapsing after several hundred iterations. It is a significant problem because for some domains we need much more iterations to obtain the acceptable quality.

The main cause of such undesirable behaviour of the $\mathcal{L}_{direction}$ (the same for $\mathcal{L}_{clip\_accross}$) loss is that it calculates the CLIP cosine distance between embeddings that do not lie in the CLIP space. Indeed, the cosine distance is a natural distance for objects that lie on a CLIP sphere but becomes less evident for vectors $\Delta T, \Delta I$ that represent the difference between clip embeddings that no longer lie on a unit sphere. Therefore, the idea behind the $\mathcal{L}_{direction}$ loss may be misleading and in practice we can observe that it still suffers from mode collapse. 

We introduce a new regularizer for improving diversity that calculates the CLIP cosine distance only between clip embeddings. We called it \emph{indomain angle consistency} loss and we define it as follows
\begin{gather}
    \mathcal{L}_{indomain-angle}(\{G^B_{d}(w_i)\}_{i=1}^n, \{G^A(w_i)\}_{i=1}^n, B, A) = \\
    = \sum\limits_{i,j}^n (\langle E_I(G^A(w_i)), E_I(G^A(w_j)) \rangle - \langle E_I(G_{d}^B(w_i)), E_I(G_{d}^B(w_j)) \rangle)^2,
\end{gather}
The idea of $\mathcal{L}_{indomain-angle}$ loss is to preserve the CLIP pairwise cosine distances between images before and after domain adaptation. We observe that this loss significantly improves the diversity of the generator $G_{d}^B$ compared to the original $L_{direction}$ or $\mathcal{L}_{clip\_accross}$ losses. 

\begin{figure}
  \includegraphics[width=\linewidth]{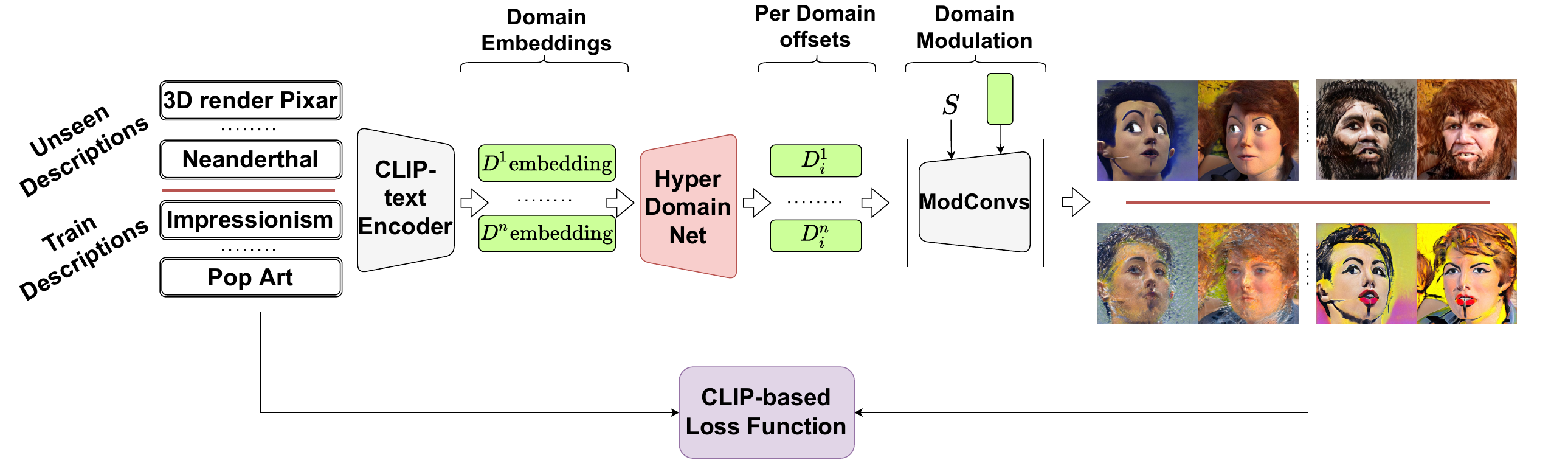}
  \caption{Detailed training process of the HyperDomainNet. On the training phase only reference descriptions are included into CLIP-guided training.}
  \label{fig:mapper_training}
\end{figure}

\subsection{Designing the HyperDomainNet for Universal Domain Adaptation}\label{sec:HDN}
The proposed domain-modulation technique allows us to reduce the number of trainable parameters which motivates us to tackle the problem of multi-domain adaption of StyleGAN2. Our aim is to train the HyperDomainNet that predicts the domain parameters given the input target domain. 
This problem can be formulated as follows. We are given a trained generator $G^A$ for a source domain $A$ and a number of target domains $B_1, \dots, B_m$ that can be represented by the single image or the text description. The aim is to learn the HyperDomainNet $D_{\varphi}(\cdot)$ that can predict the domain parameters $d_{B_i} = D_{\varphi}(B_i)$ which will be used to obtain the fine-tuned generator $G^{B_i}_{d_{B_i}}$ by the domain-modulation operation (see \Cref{sec:domain-modulation}).

In this work, we focus on the setting when the target domains $B_1, \dots, B_m$ are represented by text descriptions $t_{B_1}, \dots, t_{B_m}$. The HyperDomainNet $D_{\varphi}(\cdot)$ takes as an input the embedding of the text obtained by the CLIP encoder $E_T(\cdot)$ and outputs the domain parameters $d_{B_i} = D_{\varphi}(E_T(t_{B_i}))$. The training process is described in the \Cref{fig:mapper_training}.

To train the HyperDomainNet $D_{\varphi}(\cdot)$ we use the sum of $\mathcal{L}_{direction}$ losses for each target domains. In addition, we introduce $\mathcal{L}_{tt-direction}$ loss ("tt" stands for target-target) that is the same as $\mathcal{L}_{direction}$, but we compute it between two target domains instead of target and source. The idea is to keep away the images from different target domains in the CLIP space. We observe that without $\mathcal{L}_{tt-direction}$ loss the HyperDomainNet tends to learn the mixture of domains. 

In multi-domain adaptation setting, the regularizer $\mathcal{L}_{indomain-angle}$ becomes inefficient because during training  batch consists of samples from different domains and the number of images from one domain can be very small. Therefore, we introduce an alternative regularization $\mathcal{L}_{domain-norm}$ for the HyperDomainNet that constrains the norm of the predicted domain parameters. To be exact it equals to $\|D_{\varphi}(E_T(t_{B_i})) - 1\|^2$. 

So, the objective function of the HyperDomainNet consists of $\mathcal{L}_{direction}$, $\mathcal{L}_{tt-direction}$ and $\mathcal{L}_{domain-norm}$ losses. For more detailed description of these losses the overall optimization process, please refer to \Cref{appx:hyperdomainnet}. 


\section{Experiments}\label{sec:exps}
In this section, we provide qualitative and quantitative results of the proposed approaches. At first, we consider the text-based domain adaptation and show that our parameterization has comparable quality with the full one. Next, we tackle one-shot domain adaptation and confirm the same quantitatively and also show the importance of the $\mathcal{L}_{indomain-angle}$ loss. Finally, we solve the multi-domain adaptation problem by the proposed HyperDomainNet, show its generalization ability on unseen domains. For the detailed information about setup of the experiments please refer to \Cref{appx:setup}. 

\begin{figure}
  \includegraphics[width=\linewidth]{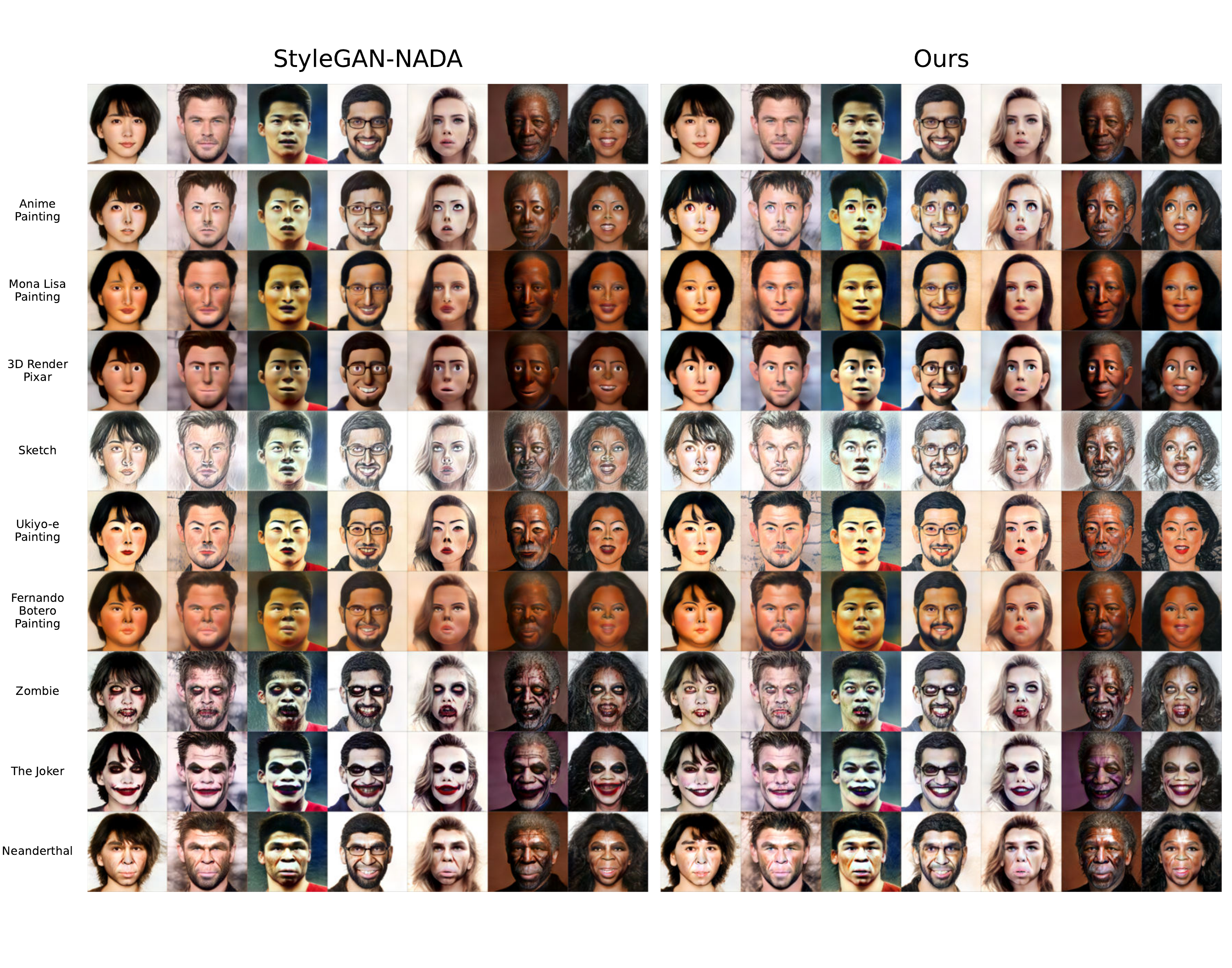}
  \vspace{-1.1cm}
  \caption{Comparison with the original StyleGAN-NADA \cite{gal2021stylegan} method (left) and its version with our parameterization.}
  \label{fig:text_based_comparison}
\end{figure}

\begin{figure}
  \includegraphics[width=\linewidth]{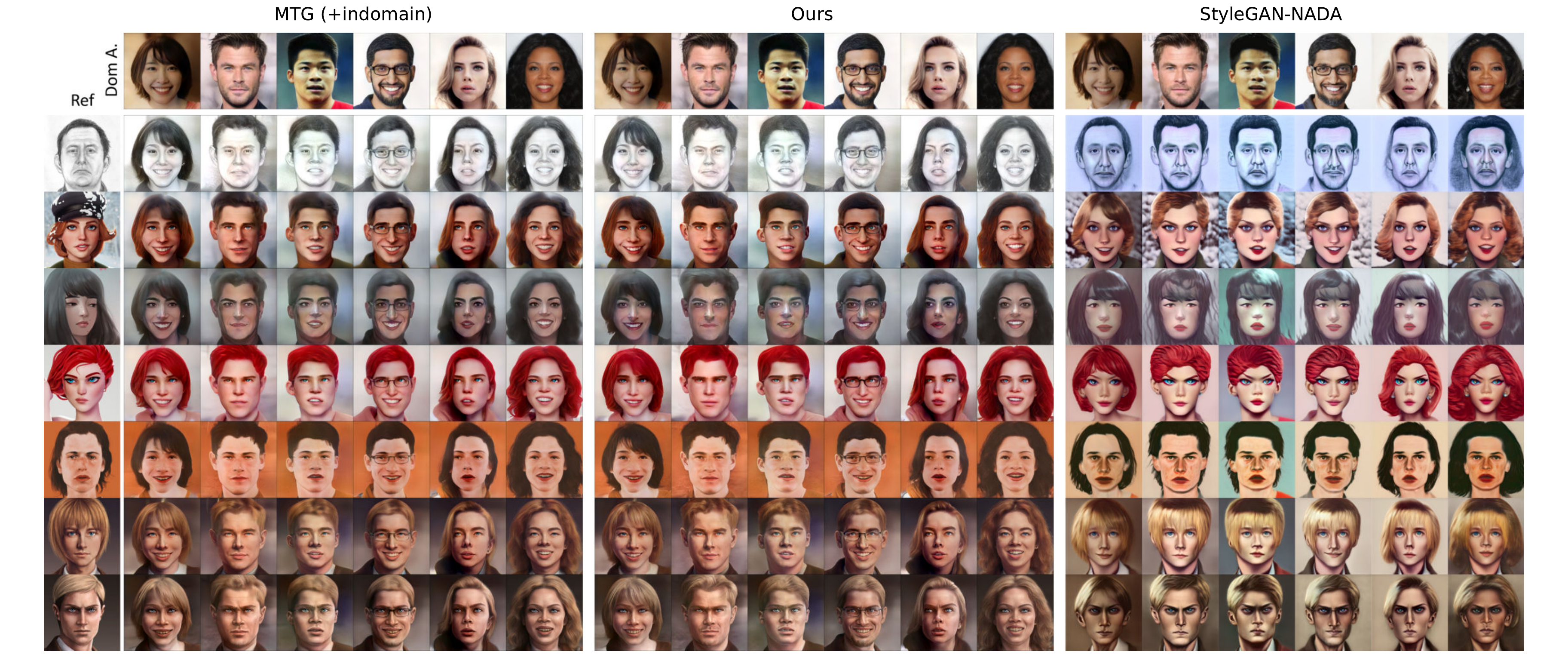}
  \caption{Comparison with one-shot domain adaptation methods. Left block is MindTheGap+indomain and right block is StyleGAN-NADA \cite{zhu2021mind}. The middle block is the MindTheGap+indomain with our parameterization.}
  \label{fig:image_based_comparison}
\end{figure}

\paragraph{Text-Based Domain Adaptation} We compare the StyleGAN-NADA \cite{gal2021stylegan} method with the proposed parameterization and the original version on a number of diverse domains. In \Cref{fig:text_based_comparison}, we see that the expressiveness of our parameterization is on par with the original StyleGAN-NADA. We observe that the domain-modulation technique allows to adapt the generator to various style and texture changes. For results on more domains please refer to \Cref{appx:text_based}. We also provide quantitative results for this setting in \Cref{appx:quant_results} which show that our parameterization has the comparable performance as the full one. 

\paragraph{One-Shot Domain Adaptation} In this part, we examine our parameterization and the indomain angle consistency loss by applying them to the MindTheGap \cite{zhu2021mind} method. We show qualitative and quantitative results and compare them with other few-shot domain adaptation methods such as StyleGAN-NADA, TargetCLIP \cite{chefer2021image} and Cross-correspondence \cite{ojha2021few} method. To assess the domain adaptation quality we use the standard metrics FID \cite{heusel2017gans}, precision and recall \cite{kynkaanniemi2019improved}. As a target domain we take the common benchmark dataset of face sketches \cite{wang2008face} that has approximately 300 samples. We consider the one-shot adaptation setting. We provide the results in \Cref{table:one-shot}. At fisrt, we see that the MindTheGap with our parameterization shows comparable results with the original version while having less trainable parameters by three orders of magnitude. While TargetCLIP has the same order of parameters as our method it shows poor adaptation quality in terms of FID and Precision metrics that indicate that it works only for in-domain editing (see also qualitative comparison in \Cref{appx:image_based} in \Cref{fig:other_methods_comparison}). 
Secondly, we examine the effectiveness of the indomain angle consistency. We show that it considerably improves FID and precision metrics for both the original MindTheGap and the one with our parameterization. 

The qualitative results are provided in \Cref{fig:image_based_comparison} for MindTheGap+indomain, MindTheGap+indomain with our parameterization ("Ours") and StyleGAN-NADA. For other methods please see \Cref{appx:image_based}. We observe that MindTheGap+indomain and our version shows comparable visual quality and outperform StyleGAN-NADA in terms of diversity and maintaining the similarity to the source image. 

Overall, we demonstrate that our parameterization is applicable to the state-of-the-art methods StyleGAN-NADA and MindTheGap and it can be further improved by the indomain angle consistency loss.

\newcommand{\mlcell}[2][p{2cm}c]{%
    \begin{tabular}[#1]{@{}c@{}}#2\end{tabular}
}
\newcolumntype{H}{>{\setbox0=\hbox\bgroup}c<{\egroup}@{}}

\begin{table*}[!b]
\centering
\caption{Evaluation of one-shot adaptation methods. Results for TargetCLIP, Cross-correspondence and StyleGAN-NADA methods are taken from \cite{zhu2021mind}. }
	\label{table:one-shot}
  \begin{tabular}{ llllc }
    \toprule
    & \multicolumn{3}{c}{\textbf{Model quality}} & \multicolumn{1}{c}{\textbf{Model complexity}} \\
    \cmidrule(lr){2-4}                  
    \cmidrule(lr){5-5}
  Model & FID & Precision & Recall & \# trainable parameters \\
    \midrule\midrule
    TargetCLIP \cite{chefer2021image} & 199.33 & 0.000 & 0.293 & 9K \\
    Cross-correspondence \cite{ojha2021few} & 158.86 & 0.001 & 0 & 30M \\
    StyleGAN-NADA \cite{gal2021stylegan} & 124.55 & 0.118 & 0 & 24M \\
    MindTheGap \cite{zhu2021mind} & 78.35 & 0.326 & 0.017 & 24M \\
    \midrule
    MindTheGap (our param.) & 79.83 & 0.452 & 0.017 & 6k \\
    \midrule
    MindTheGap+indomain & 71.46 & 0.503 & 0.014 & 24M \\
    MindTheGap+indomain (our param.) & 72.71 & 0.472 & 0.028 & 6k \\
    \midrule
    \bottomrule
  \end{tabular}
  \vspace{-0.5cm}
\end{table*}


\paragraph{Multi-Domain Adaptation} Now we consider the multi-domain adaptation problem. We apply the HyperDomainNet in two different scenarios: (i) training on fixed number of domains, (ii) training on potentially arbitrary number of domains. The first scenario is simple, we train the HyperDomainNet on 20 different domains such as "Anime Painting", "Pixar", etc. (for the full list of domains please refer to \Cref{sec:fixed_number}). The second scheme is more complicated. We fix large number of domains (several hundreds) and calculate its CLIP embeddings. During training we sample new embeddings from the convex hull of the initial ones and use them in the optimization process (see \Cref{fig:mapper_training}). This technique allows us to generalize to unseen domains. For more details about both scenarios please refer to \Cref{appx:hyperdomainnet}. 

\begin{figure}
  \includegraphics[width=\linewidth]{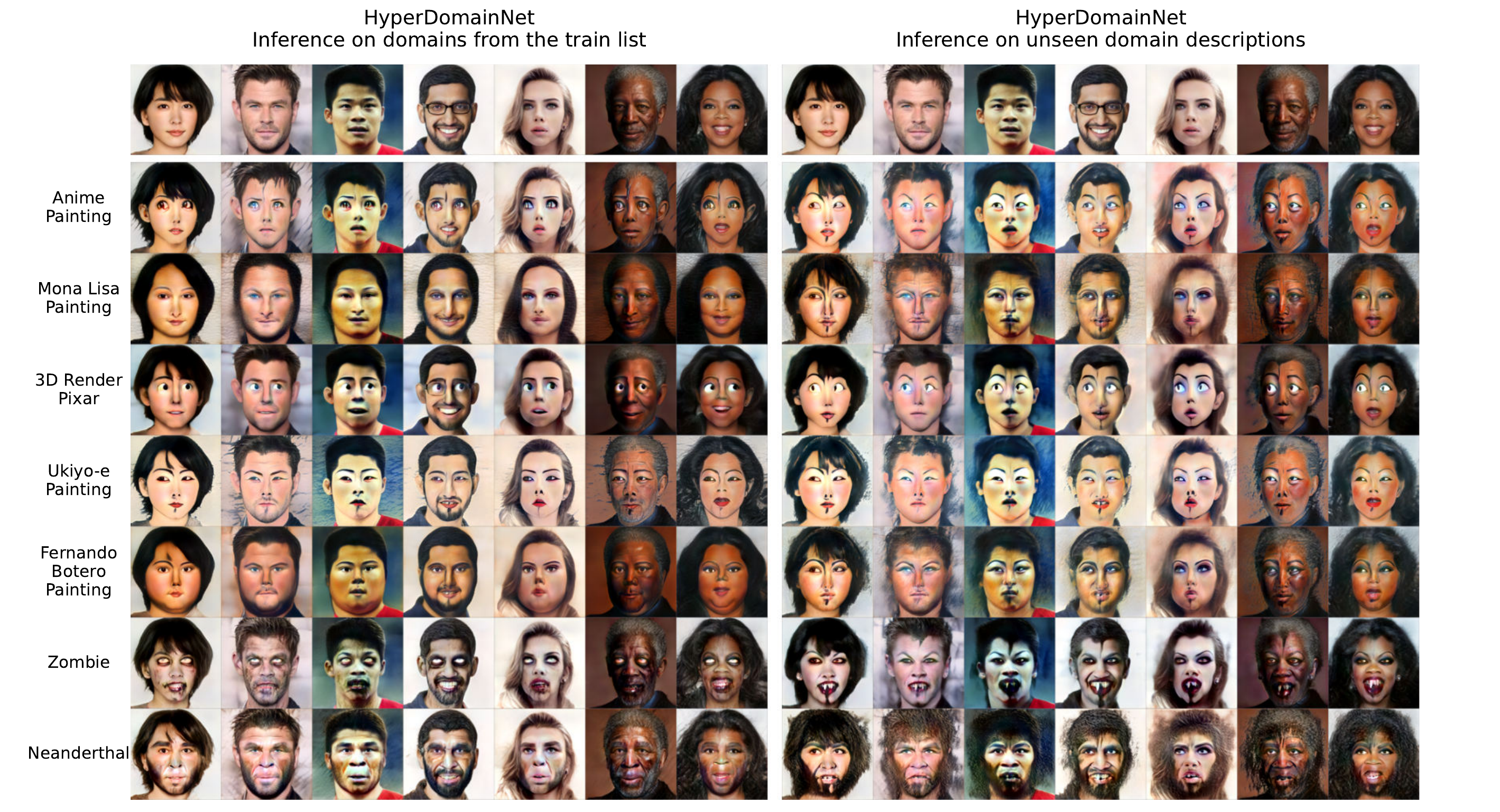}
  \caption{Comparison of training setups. The top row represents the real images embedded into StyleGAN2 latent space which latents are then used for HyperDomainNet inference. The left block represents results obtained from text-descriptions presented in the train list. The right block represents results of HyperDomainNet inference on unseen text-descriptions.}
  \label{fig:mapper_comparison}
\end{figure}

The results of the HyperDomainNet for both scenarios are provided in \Cref{fig:mapper_comparison}. The left part is results for the first setting, the right one is results for the unseen domains in the second scheme. For more domains and generated images please refer to \Cref{appx:hyperdomainnet}. We see that in the first scenario the HyperDomainNet shows results comparable to the case when we train separate models for each domain (see \Cref{fig:text_based_comparison}). It shows that the proposed optimization process for the HyperDomainNet is effective. The results for the second scenario looks promising. We can observe that the HyperDomainNet has learnt very diverse domains and shows sensible adaptation results for unseen ones. 

We also provide an ablation study on the loss terms we use for training of the HyperDomainNet in \Cref{sec:ablation_study}. It demonstrates quantitatively and qualitatively that the proposed losses are essential for the effective training of the HyperDomainNet in the setting of the multi-domain adaptation problem.

\section{Conclusion}\label{sec:conclusion}
We propose a novel domain-modulation technique that allows us to considerably reduce the number of trainable parameters during domain adaptation of StyleGAN2. In particular, instead of fine-tuning almost all 30 million weights of the StyleGAN2 we optimize only 6 thousand-dimensional domain vector. We successfully apply this technique to the state-of-the-art text-based and image-based domain adaptation methods. We show quantitatively and qualitatively that it can achieve the same quality as optimizing all weights of the StyleGAN2.

To deal with the mode collapsing problem of the domain adaptation methods we introduce a new indomain angle consistency loss $\mathcal{L}_{indomain-angle}$ that preserves the CLIP pairwise cosine distances between images before and after domain adaptation. We demonstrate that it improves the diversity of the fine-tuned generator both for text-based and one-shot domain adaptation. 

We also consider the problem of multi-domain adaptation of StyleGAN2 when we aim to adapt to several domains simultaneously. Before our proposed parameterization it was infeasible because we should predict all weights of StyleGAN2 for each domain. Thanks to our efficient parameterization we propose HyperDomainNet that predicts the 6 thousand-dimensional domain vector $d$ for the StyleGAN2 given the input domain. 
We empirically show that it can be trained to 20 domains successfully which is the first time when the StyleGAN2 was adapted to several domains simultaneously. We also train the HyperDomainNet for the large number of domains (more than two hundred) with applying different augmentations to the domain descriptions (see details in \Cref{appx:hyperdomainnet}). We demonstrate in practice that in such setting the HyperDomainNet can generalize to unseen domains.  

\paragraph{Limitations and societal impact} The main limitation of our approach is that it is not applicable for the cases when target domains are very far from the source one. In such setting, we cannot limit the parameter space, so we should use the full parameterization. 

The potential negative societal impacts of domain adaptation of GANs and generally training of GANs include different forms of disinformation, e.g. deepfakes of celebrities or senior officials, fake avatars in social platforms. However, it is the issue of the whole field and this work does not amplify this impact. 


\section{Acknowledgments and Disclosure of Funding}
The publication was supported by the grant for research centers in the field of AI provided by the Analytical Center for the Government of the Russian Federation (ACRF) in accordance with the agreement on the provision of subsidies (identifier of the agreement 000000D730321P5Q0002) and the agreement with HSE University No. 70-2021-00139. Additional revenues of the authors for the last three years: laboratory sponsorship by Samsung Research, Samsung Electronics and Huawei Technologies; Institute for Information Transmission Problems, Russian Academy of Science. 

\newpage

{\small
\bibliographystyle{plain}
\bibliography{ref}
}

\newpage

\newpage
\appendix

\section{Appendix}

\subsection{Setup of the Experiments}\label{appx:setup}
\subsubsection{Implementation Details}
We implement our experiments using PyTorch\footnote{\hyperlink{https://pytorch.org}{https://pytorch.org}} deep learning framework. For StyleGAN2 \cite{karras2020analyzing} architecture we use the popular PyTorch implementation \footnote{\hyperlink{https://github.com/rosinality/stylegan2- pytorch}{https://github.com/rosinality/stylegan2- pytorch}}. We attach all source code that reproduces our experiments as a part of the supplementary material. We also provide configuration files to run each experiment.

\subsubsection{Datasets}
We use source StyleGAN2 models pretrained on the following datasets: (i) Flickr-Faces-HQ (FFHQ) \cite{karras2019style}, (ii) LSUN Church, (iii) LSUN Cars, and (iv) LSUN Cats \cite{yu2015lsun}. As target domains we mainly use the text descriptions from \cite{gal2021stylegan} and style images from \cite{zhu2021mind}. For quantitative comparison with other methods we use face sketches \cite{wang2008face} as the standard dataset for domain adaptation. 

\subsubsection{Licenses and Data Privacy}
Tables \ref{tbl:model_licenses}, \ref{tbl:data_licenses} provide sources and licenses of the models and datasets we used in our work. 

\begin{table}[h]
    \centering
    \caption{Models used in our work, their sources and licenses.}
    \begin{tabular}{l c c}
        Model & Source & License \\ \hline
        StyleGAN2 & \cite{karras2020analyzing} & \nvsrc \\ 
        pSp & \cite{richardson2021encoding} & \mitlic \\
        e4e & \cite{tov2021designing} & \mitlic \\
        ReStyle & \cite{alaluf2021restyle} & \mitlic \\
        StyleCLIP & \cite{patashnik2021styleclip} & \mitlic \\
        CLIP & \cite{radford2021learning} & \mitlic \\
        StyleGAN2-pytorch & \cite{rosinalitySG2} & \mitlic \\
        StyleGAN-ADA & \cite{karras2020training} & \href{https://nvlabs.github.io/stylegan2-ada-pytorch/license.html}{Nvidia Source Code License} \\
        Cross-correspondence & \cite{ojha2021few} & \adblic
    \end{tabular}\label{tbl:model_licenses}
\end{table}

\begin{table}[h]
    \centering
    \caption{Datasets used in our work, their sources and licenses.}
    \begin{tabular}{l c c}
        Dataset & Source & License \\ \hline
        FFHQ & \cite{karras2019style} & \ccbyncsa \footnote{Individual images under different licenses. See \url{https://github.com/NVlabs/ffhq-dataset}}\\
        LSUN & \cite{yu2015lsun} & No License \\ 
        Sketches & \cite{ojha2021few} & \adblic
    \end{tabular}
    \label{tbl:data_licenses}
\end{table}

\subsubsection{Total Amount of Compute Resources}
We run our experiments on Tesla A100 GPUs. We used approximately 12000 GPU hours to obtain the reported results and for intermediate experiments. 

\subsection{Training of the HyperDomainNet (HDN)}\label{appx:hyperdomainnet}
\subsubsection{Training Losses}
\label{appx:train_losses}
As we describe in \Cref{sec:HDN} we train HDN $D_{\varphi}(\cdot)$ using three losses $\mathcal{L}_{direction}$, $\mathcal{L}_{tt-direction}$, and $\mathcal{L}_{domain-norm}$. Each loss is defined as follows:
\begin{gather}
    \mathcal{L}_{direction}(G^{B_i}_{d_{B_i}}(w), G^{A}(w), B_i, A) = 1 - \dfrac{\Delta I(G^{B_i}_{d_{B_i}}(w), G^{A}(w)) \cdot \Delta T(B_i, A)}{|\Delta I(G^{B_i}_{d_{B_i}}(w), G^{A}(w))| |\Delta T(B_i, A)|}, \\
    \mathcal{L}_{tt-direction}(G^{B_i}_{d_{B_i}}(w), G^{B_j}_{d_{B_j}}(w), B_i, B_j) = 1 - \dfrac{\Delta I(G^{B_i}_{d_{B_i}}(w), G^{B_j}_{d_{B_j}}(w)) \cdot \Delta T(B_i, B_j)}{|\Delta I(G^{B_i}_{d_{B_i}}(w), G^{B_j}_{d_{B_j}}(w))| |\Delta T(B_i, B_j)|}, \\
    \mathcal{L}_{domain-norm}(D_{\varphi}, B_i) = \|D_{\varphi}(E_T(t_{B_i})) - 1\|^2
\end{gather}

Then the overall training loss for the HDN $D_{\varphi}(\cdot)$ is
\begin{gather}
    \mathcal{L}(\varphi) = \lambda_{direction} \sum\limits_{\substack{i=1}}^m \mathcal{L}_{direction}(G^{B_i}_{D_{\varphi}(E_T(t_{B_i}))}(w), G^{A}(w), B_i, A) \; + \nonumber \\
    \\ + \; \lambda_{tt-direction} \sum\limits_{\substack{i\neq j}}^m \mathcal{L}_{tt-direction}(G^{B_i}_{D_{\varphi}(E_T(t_{B_i}))}(w), G^{B_j}_{D_{\varphi}(E_T(t_{B_j}))}(w), B_i, B_j) \; + \nonumber \\
    + \; \lambda_{domain-norm}\sum\limits_{i=1}^m \mathcal{L}_{domain-norm}(D_{\varphi}, B_i) \label{eq:hdn_loss}
\end{gather}

\subsubsection{Architecture of the HDN}
\begin{figure}[!h]
\centering
  \includegraphics[width=0.8\textwidth]{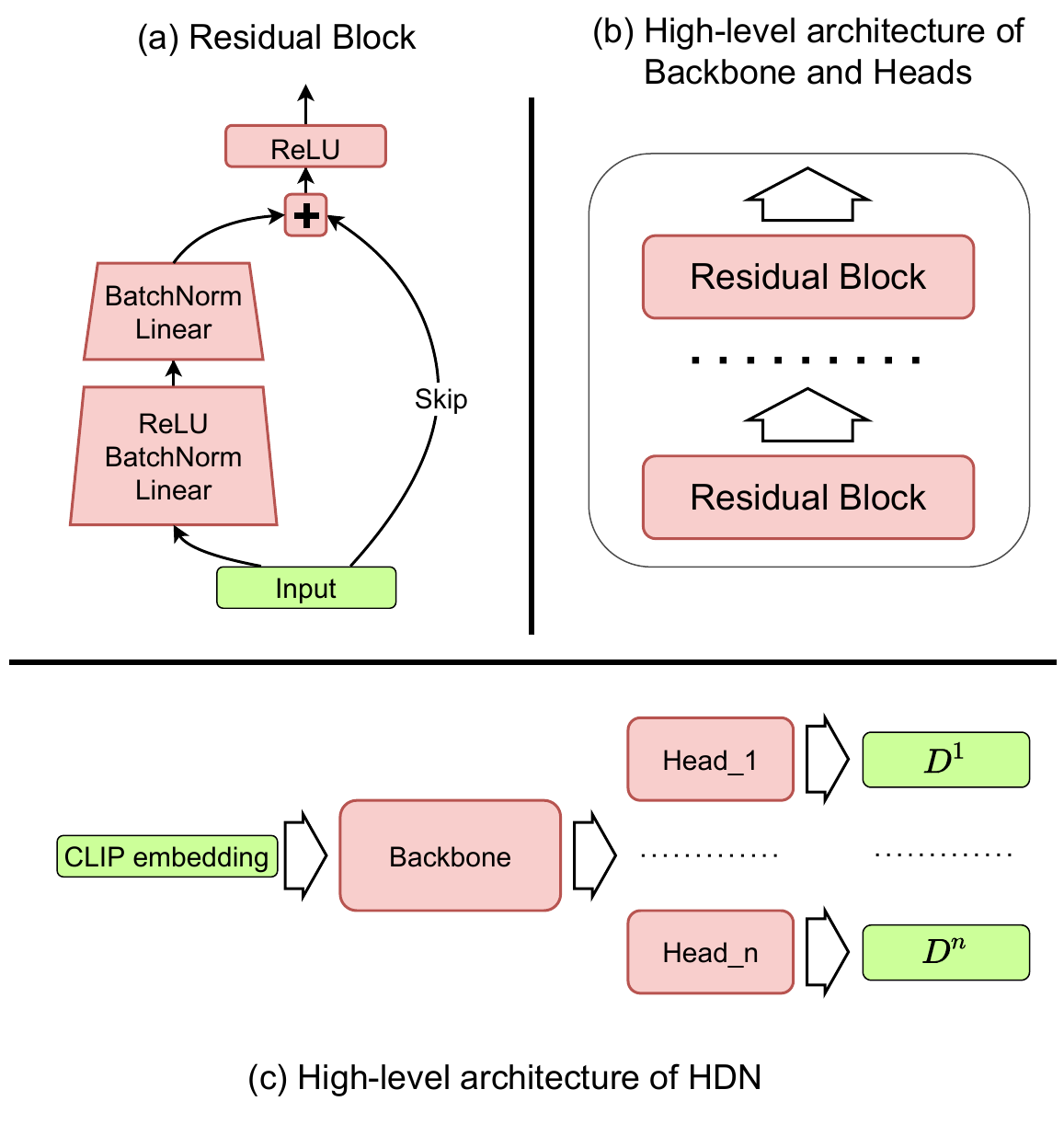}
  \caption{Detailed HDN architecture diagram. (a) - base residual block which is included into backbone and head parts of the HDN. (b) - the detailed backbone and head architecture, each module use the same sequence of ResBlocks. (c) - the detailed architecture of the HDN with data flow.}
  \label{fig:mapper_architecture}
\end{figure}

We use the standard ResNet-like architecture for the HDN. It has the backbone part which has 10 ResBlocks and the part that consists of 17 heads. The number of heads equals the number of StyleGAN2 layers in the synthesis network $G_{sys}$. Each head has 5 ResBlocks and outputs the domain vector for the corresponding StyleGAN2 layer. We illustrate the overall architecture of the HDN in Figure \ref{fig:mapper_architecture}. It has 43M parameters. We use the same architecture for all experiments. 


\subsubsection{Inference Time}
The inference time of the HDN network on 1 Tesla A100 GPU is almost the same as the one forward pass through StyleGAN2 generator which works in 0.02 seconds.

\subsubsection{Training on Fixed Number of Domains}\label{sec:fixed_number}
For training the HDN on fixed number of domains we use the loss function from \Cref{eq:hdn_loss}. As training target domains we take the following 20 domains (we provide in the format "the target domain - the corresponding source domain"):
\begin{enumerate}
    \item Anime Painting - Photo
    \item Impressionism Painting - Photo
    \item Mona Lisa Painting - Photo
    \item 3D Render in the Style of Pixar - Photo
    \item Painting in the Style of Edvard Munch - Photo
    \item Cubism Painting - Photo
    \item Sketch - Photo
    \item Dali Painting - Photo
    \item Fernando Botero Painting - Photo
    \item A painting in Ukiyo-e style - Photo
    \item Tolkien Elf - Human
    \item Neanderthal - Human
    \item The Shrek - Human
    \item Zombie - Human
    \item The Hulk - Human
    \item The Thanos - Human
    \item Werewolf - Human
    \item Nicolas Cage - Human
    \item The Joker - Human
    \item Mark Zuckerberg - Human
\end{enumerate}

\paragraph{Hyperparameters}
For 20 descriptions in training list following setup is considered. The HDN trained for 1000 number of iterations. Batch size 24 is used. Weights of the terms from \Cref{eq:hdn_loss} as follows: $\lambda_{direction} = 1.0$, $\lambda_{tt-direction} = 0.4$, $\lambda_{domain-norm} = 0.8$. We use two Vision-Transformer based CLIP models, "ViT-B/32" and "ViT-B/16". To optimize HDN we use an ADAM Optimizer with betas$= (0.9, 0.999)$, learning rate$=5e-5$, weight decay$=0$.

\subsubsection{Training Time}
The training time of the HDN on 20 domains for 1000 iterations on single Tesla A100 GPUs takes about 2 hours. 

\subsubsection{Ablation Study on the Loss Terms}
\label{sec:ablation_study}
We perform both the quantitative and qualitative ablation study on the domain-norm and tt-direction loss terms that are defined in \Cref{appx:train_losses}. 

\begin{figure}[!h]
  \includegraphics[width=\textwidth]{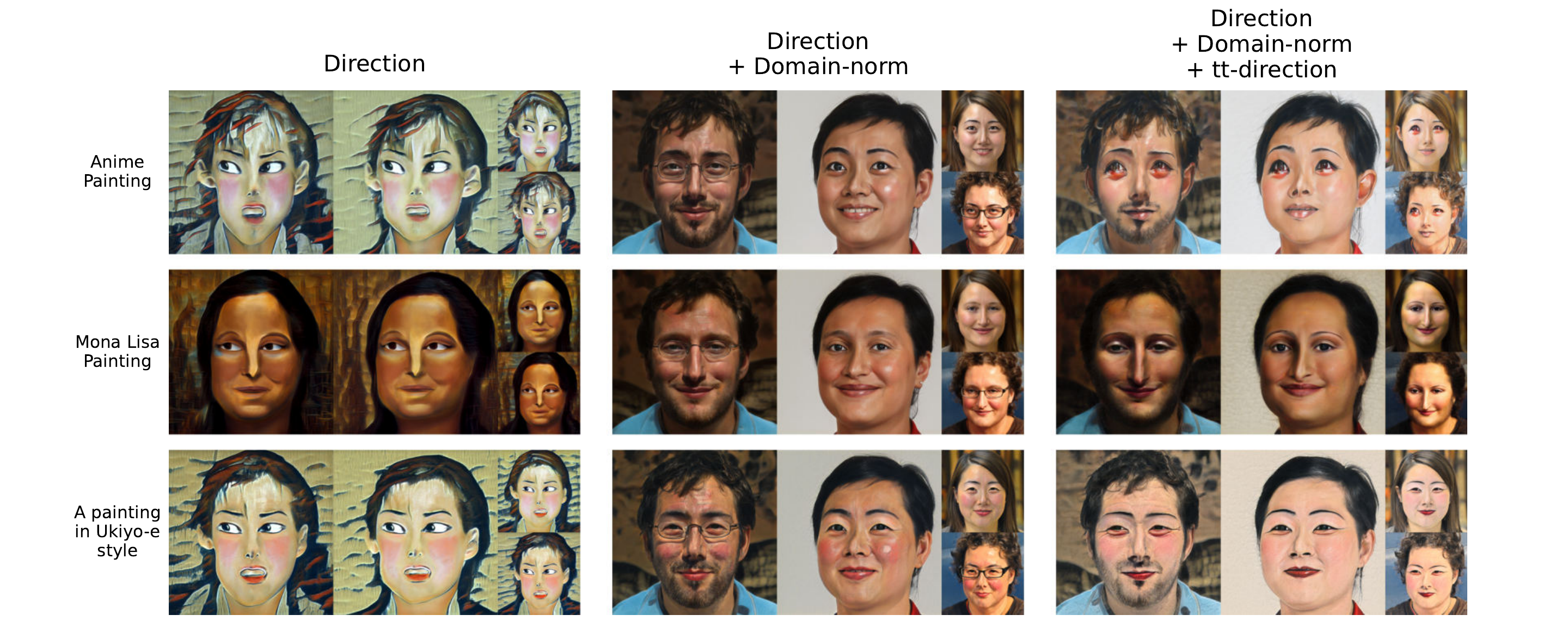}
  \caption{Ablation study on the loss terms of the HyperDomainNet.}
  \label{fig:ablation_study}
\end{figure}

For the qualitative analysis we consider three domains (Anime Painting, Mona Lisa Painting, A painting in Ukiyo-e style) for the HyperDomainNet that was trained on 20 different domains (see the full list in \Cref{sec:fixed_number}). We provide the visual comparison for these domains with respect to the using loss terms in the training loss of the HyperDomainNet (see \Cref{fig:ablation_study}). We can see that without additional loss terms the model considerably collapses within each domain. After adding domain-norm it solves the problem of collapsing within domains but it starts mix domains with each other, so we obtain the same style for different text descriptions. And after using tt-direction loss eventually allows us to train the HyperDomainNet efficiently on these domains without collapsing. 

For the quantitative results we use the metrics Quality and Diversity that were introduced in \Cref{appx:quant_results}. The results are provided in \Cref{table:ablation_study}. We see that the initial model without loss terms obtains good Quality but very low Diversity. The domain-norm significantly improves the diversity in the cost of degrading the Quality. The tt-direction provides a good balance between these two metrics which we also we qualitatively in \Cref{fig:ablation_study}.  

\begin{table*}[!t]
\centering
\caption{Ablation study on the loss terms of the HyperDomainNet.}
	\label{table:ablation_study}
  \begin{tabular}{ lll }
    \toprule
  Model & Quality & Diversity\\
    \midrule\midrule
    Anime Painting \\
    \midrule
    Multi-Domain & 0.271 & 0.128 \\
    Multi-Domain+domain\_norm & 0.210 & 0.338 \\
    Multi-Domain+domain\_norm+tt\_direction & 0.260 & 0.256 \\
    \midrule
    Zombie \\
    \midrule
    Multi-Domain & 0.254 & 0.079 \\
    Multi-Domain+domain\_norm & 0.246 & 0.203 \\
    Multi-Domain+domain\_norm+tt\_direction & 0.258 & 0.191 \\
    \midrule
    Across ten domains \\
    \midrule
    Multi-Domain & 0.275 $\pm$ 0.035 & 0.099 $\pm$ 0.026 \\
    Multi-Domain+domain\_norm & 0.218 $\pm$ 0.026 & 0.306 $\pm$ 0.040 \\
    Multi-Domain+domain\_norm+tt\_direction & 0.247 $\pm$ 0.026 & 0.250 $\pm$ 0.041 \\
    \midrule
    \bottomrule
  \end{tabular}
  \vspace{-0.05cm}
\end{table*}

\begin{figure}[!h]
  \includegraphics[width=\textwidth]{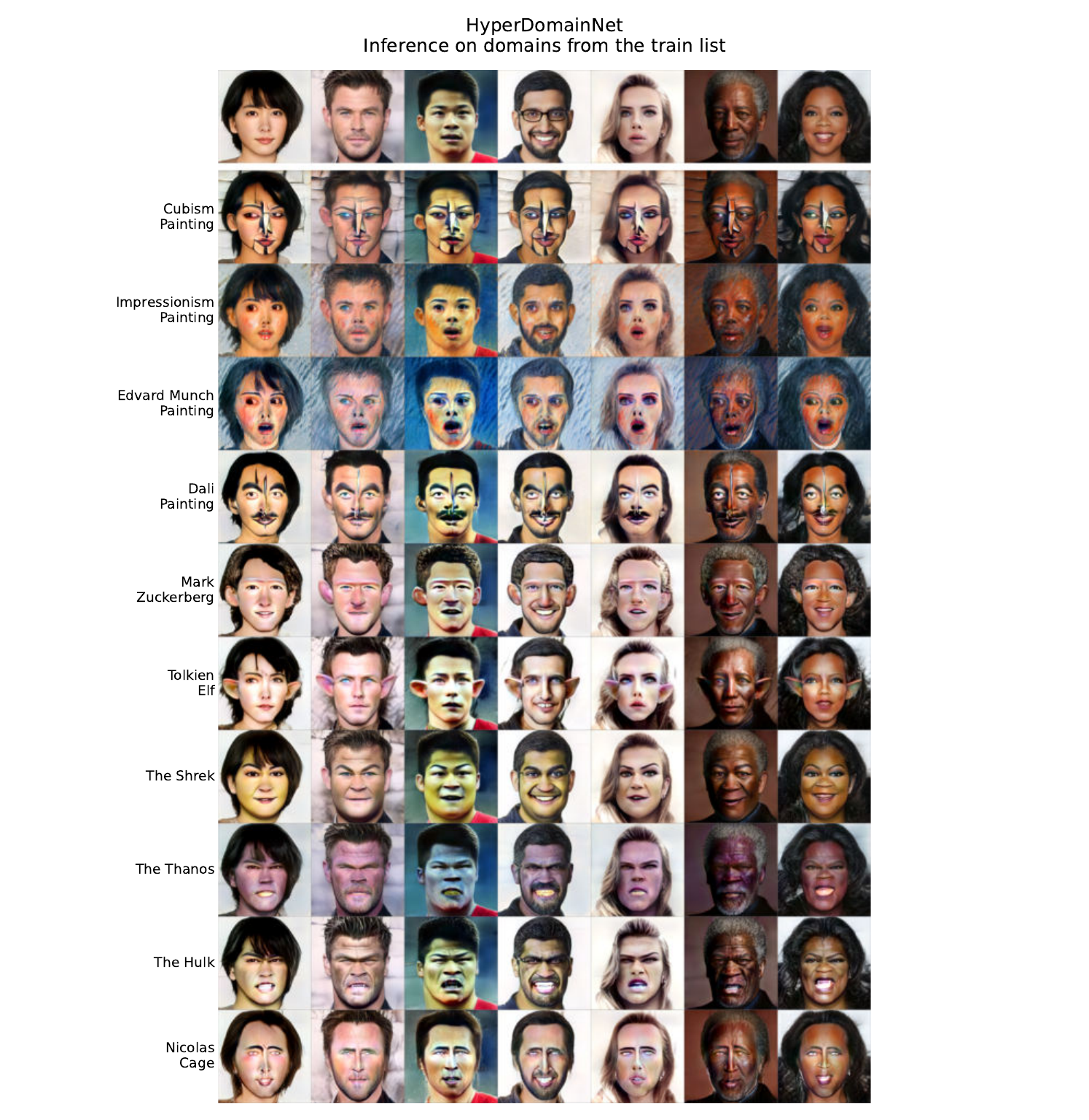}
  \caption{Other domains which were included into train description list from left block of \Cref{fig:mapper_comparison}.}
  \label{fig:supp_20_mapper}
\end{figure}

\paragraph{Additional Samples}
We show results for the first 10 domains in \Cref{fig:mapper_comparison}. The next 10 domains we provide in \Cref{fig:supp_20_mapper}.

\FloatBarrier

\subsubsection{Training on Potentially Arbitrary Number of Domains}
Improving the generalization ability of the HDN is challenging because it tends to overfit on training domains and for unseen ones it predicts domains that are very close to train ones. One way to tackle this problem is to considerably extend the training set. For this purpose, we use three techniques: (i) generating many training domains by taking combinations of different ones; (ii) sampling CLIP embeddings from convex hull of the initial training embeddings; (iii) resample initial CLIP embeddings given cosine similarity. We discuss each technique further.

\paragraph{Generating Training Domains by Taking Combinations}
We can describe domains by indicating different properties of the image such as image style (e.g. Impressionism, Pop Art), image type (e.g. Photo, Painting), artist style (e.g. Modigliani). Also we can construct new domains by taking combinations of these properties (e.g. Modigliani Painting, Impressionism Photo). So, we take 32 image styles, 13 image types and 7 artists and by taking all combinations of these properties we come up with 1040 domains. We provide the full list of properties we use in our training:
\begin{itemize}
    \item image styles: 'Pop Art', 'Impressionism', 'Renaissance', 'Abstract', 'Vintage', 'Antiquity', 'Cubism', 'Disney', 'Chinese', 'Japanese', 'Spanish', 'Italian', 'Dutch', 'German', 'Surreal', 'WaltDisney', 'DreamWorks', 'Modern', 'Realism', 'Starry Night', 'Old-timey', 'Pencil', 'Gouache', 'Acrylic', 'Watercolor', 'Oil', 'Black', 'Blue', 'Charcoal', 'Manga', 'Kodomo';
    \item image types: 'Portrait', 'Image', 'Photo', 'Painting', 'Graffiti', 'Photograph', 'Cartoon', 'Stereo View', 'Drawing', 'Graphics', 'Mosaic', 'Caricature', 'Animation';
    \item artists: 'Raphael', 'Salvaror Dali', 'Edvard Munch', 'Modigliani', 'Van Gogh', 'Claude Monet', 'Leonardo Da Vinci'
\end{itemize}

The algorithm of generating combinations is
\begin{figure}[!h]
  \includegraphics[width=0.9\textwidth]{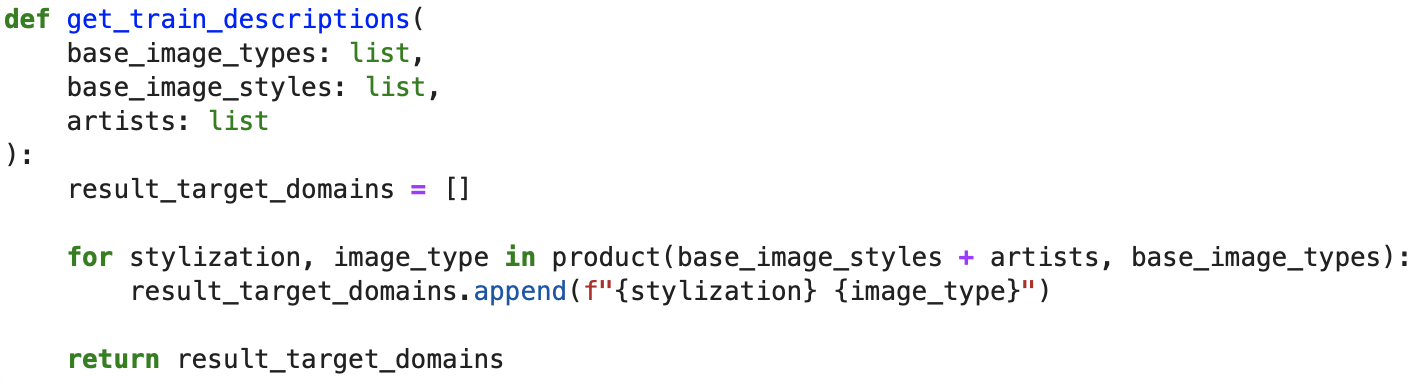}
  \caption{The algorithm of generating combinations implemented on \textbf{Python}.}
  \label{fig:constructor_python}
\end{figure}
\FloatBarrier

\paragraph{Generating CLIP Embeddings from Convex Hull}
In the usual training of the HDN as in \Cref{sec:fixed_number} we use the CLIP embeddings of the target domains $t_{B_1} = E_T(B_1), \dots, t_{B_m} = E_T(B_m)$. To cover more CLIP space we propose to use new embeddings $t'_{B_1}, \dots, t'_{B_m}$ from the convex hull of the initial ones:
\begin{gather}
    t'_{B_i} = \sum\limits_{j=1}^m \alpha_j t_{B_j}, \; i = 1, \dots, m, \\
    \text{where } \alpha_1, \dots, \alpha_m \sim \text{Dir}(\beta) \text{ (Dirichlet distribution)}, \; \sum\limits_{i=1}^m \alpha_i = 1, \; \alpha_i \geqslant 0, i = 1, \dots, m. 
\end{gather}
We use $\beta = \dfrac{1}{\textit{batch size}}$.

\paragraph{Resampling Initial CLIP Embeddings Given Cosine Similarity}
To further extend the CLIP space we cover during training of the HDN we resample initial CLIP embeddings of the target domains $t_{B_1}, \dots, t_{B_m}$ constrained to the cosine similarity. So, before generating from convex hull we replace the initial embeddings by new ones $\hat{t}_{B_1}, \dots, \hat{t}_{B_m}$ such that $\cos(t_{B_1}, \hat{t}_{B_1}) = \gamma$. To obtain these embeddings we use the following operation:
\begin{gather}
    \hat{t}_{B_i} = \text{resample}(t_{B_i}), \; i = 1, \dots, m, \\
    \text{where } \text{resample}(t_{B_i}) = t_{B_i}\cdot \cos\gamma + \text{norm}(v - \vectorproj[t_{B_i}]{v})\cdot\sin\gamma, \\
    v \sim \mathcal{N}(v | 0, \mathbf{I}), \quad \text{norm}(u) = \dfrac{u}{||v||_2} 
\end{gather}
It allows us to cover the part of the CLIP space outside of the initial convex hull. We observe that it improves the generalization ability of the HDN. 

\paragraph{Hyperparameters}
We train the HDN for 10000 number of iterations. We use batch size of 96. We set weights of the terms from \Cref{eq:hdn_loss} as follows: $\lambda_{direction} = 1.0$, $\lambda_{tt-direction} = 0.4$, $\lambda_{domain-norm} = 0.8$. We use two Vision-Transformer based CLIP models, "ViT-B/32" and "ViT-B/16". To optimize HDN we use an ADAM Optimizer with betas$= (0.9, 0.999)$, learning rate$=5e-5$, weight decay$=0$.


\paragraph{Training Time}
The training time of the HDN for 10000 iterations on 4 Tesla A100 GPUs takes about 50 hours. 

\begin{figure}[!h]
  \includegraphics[width=\textwidth]{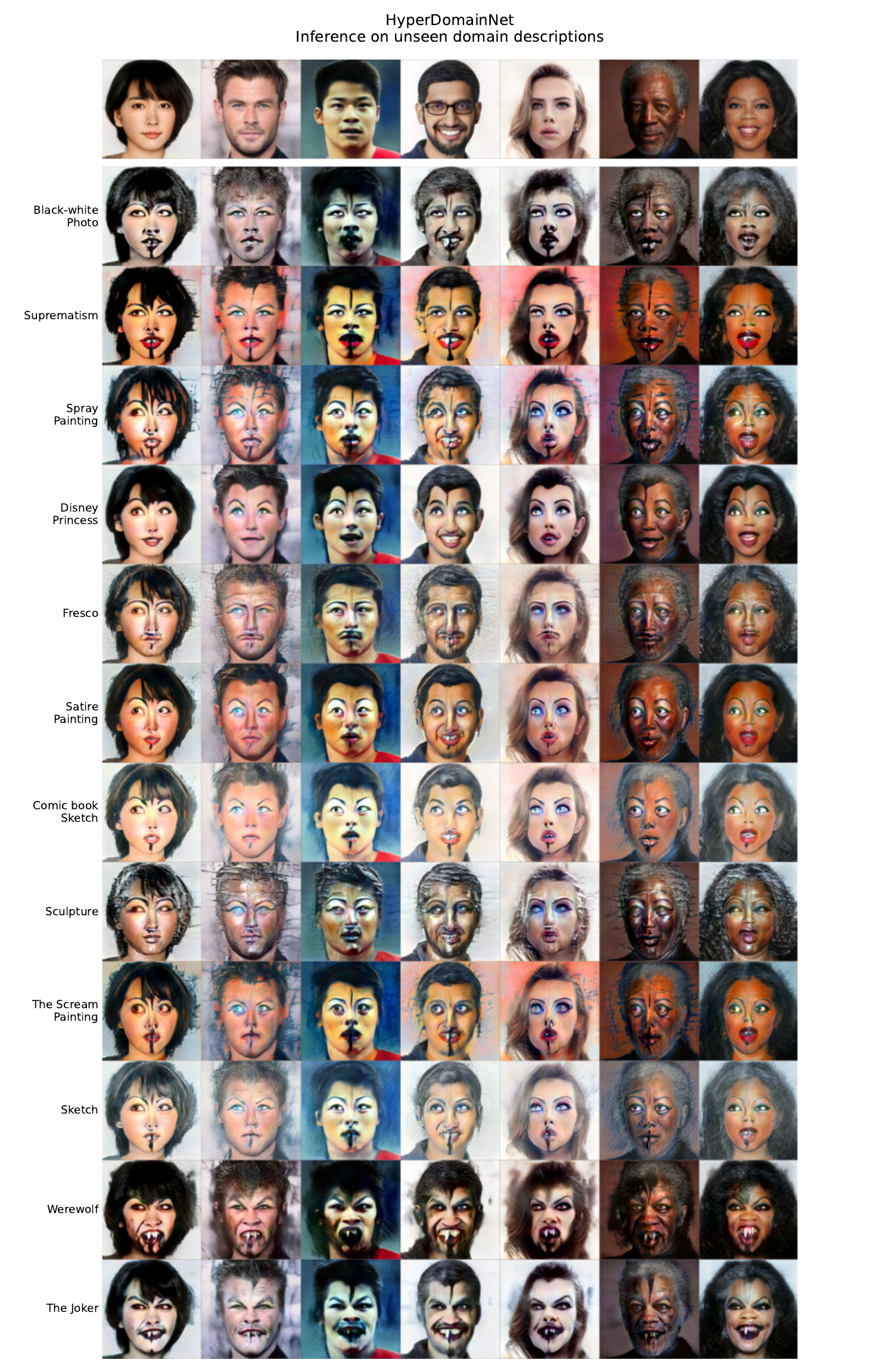}
  \caption{Other visual results for descriptions which were not included into training list during HDN training.}
  \label{fig:supp_unseen_mapper}
\end{figure}

\paragraph{Additional Samples} 
Additional samples of unseen domains for the HDN is demonstrated in \Cref{fig:supp_unseen_mapper}.

\FloatBarrier

\subsection{Results for Text-Based Domain Adaptation}\label{appx:text_based}

\subsubsection{Hyperparameters}
StyleGAN-NADA with our parameterization trained for 600 iterations with batch size of 4. Style mixing probability is set to $0.9$, the weight of the $\mathcal{L}_{direction}$ is $1.0$ and $\mathcal{L}_{indomain-angle}$ is $0.5$ and ADAM optimizer with betas$= (0., 0.999)$, learning rate$=0.002$, weight decay$=0$.

For the original StyleGAN-NADA \cite{gal2021stylegan} number of iterations is decreased to 200 because for more iterations it starts to collapse. 

"ViT-B/32" and "ViT-B/16" CLIP Vision-Transformed models used in all setups.

\subsubsection{Training and Inference Time}
The training of the one target domain for 600 iterations on a single Tesla A100 GPU takes about 15 minutes on batch size 4. 

The inference time consists of two parts. The first one is the embedding process of the real image which takes 0.23 seconds using ReStyle \cite{alaluf2021restyle}. The second part is the forward pass through adapted GAN generator which works in 0.02 seconds. 

\subsubsection{Quantitative Results}
\label{appx:quant_results}
We provide the quantitative comparison for the text-based domain adaptation by evaluating the "Quality" and "Diversity" metrics in a straightforward way. 

As the “Quality” metric we estimate how close the adapted images to the text description of the target domain. That is we compute the mean cosine similarity between image CLIP embeddings and the embedding of the text description:

\begin{gather}
    \text{Quality} = \dfrac{1}{n} \sum\limits_{i=1}^n \langle E_T(\text{target\_text}), E_I(I_i) \rangle, \text{ where} \\
n \text{ - number of the generated adapted images (we use 1000)}, \nonumber \\
E_T \text{ - text CLIP encoder}, \nonumber \\
E_I \text{ - image CLIP encoder}, \nonumber \\
I_1, \dots, I_n \text{ - generated adapted images}.\nonumber 
\end{gather}
As $E_I$ encoder we use only ViT-L/14 image encoder that is not applied during training (in the training we use ViT-B/16, ViT-B/32 image encoders). 

As the “Diversity” metric we estimate the mean pairwise cosine distance between all adapted images:
\begin{gather}
    \text{Diversity} = \dfrac{2}{n(n-1)} \sum\limits_{i < j}^n (1 - \langle E_I(I_i), E_I(I_j) \rangle), \text{ where} \\
n \text{ - number of the generated adapted images (we use 1000)}, \nonumber \\
E_I \text{ - image CLIP encoder}, \nonumber \\
I_1, \dots, I_n \text{ - generated adapted images}.\nonumber 
\end{gather}

We compute these two metrics for the ten text domains: Anime Painting, Mona Lisa Painting, 3D Render Pixar, Sketch, Ukiyo-e Painting, Fernando Botero Painting, Werewolf, Zombie, The Joker, Neanderthal. We separately report metrics for two domains Anime Painting and Zombie to better reflect the metrics behaviour. Also we report the overall metrics across all nine domains. The results are provided in \Cref{table:text-quant}. 

From these results we see that our model performs comparably with the StyleGAN-NADA with respect to Quality while having better Diversity. Also we can observe that the indomain angle loss significantly improves the Diversity for both models StyleGAN-NADA and Ours while lightly decreases the Quality. 

For the multi-domain adaptation model we see that it has lower diversity than StyleGAN-NADA and Ours and comparable Quality while being adapted to all these domains simultaneously. 

\begin{figure}[!h]
  \includegraphics[width=\textwidth]{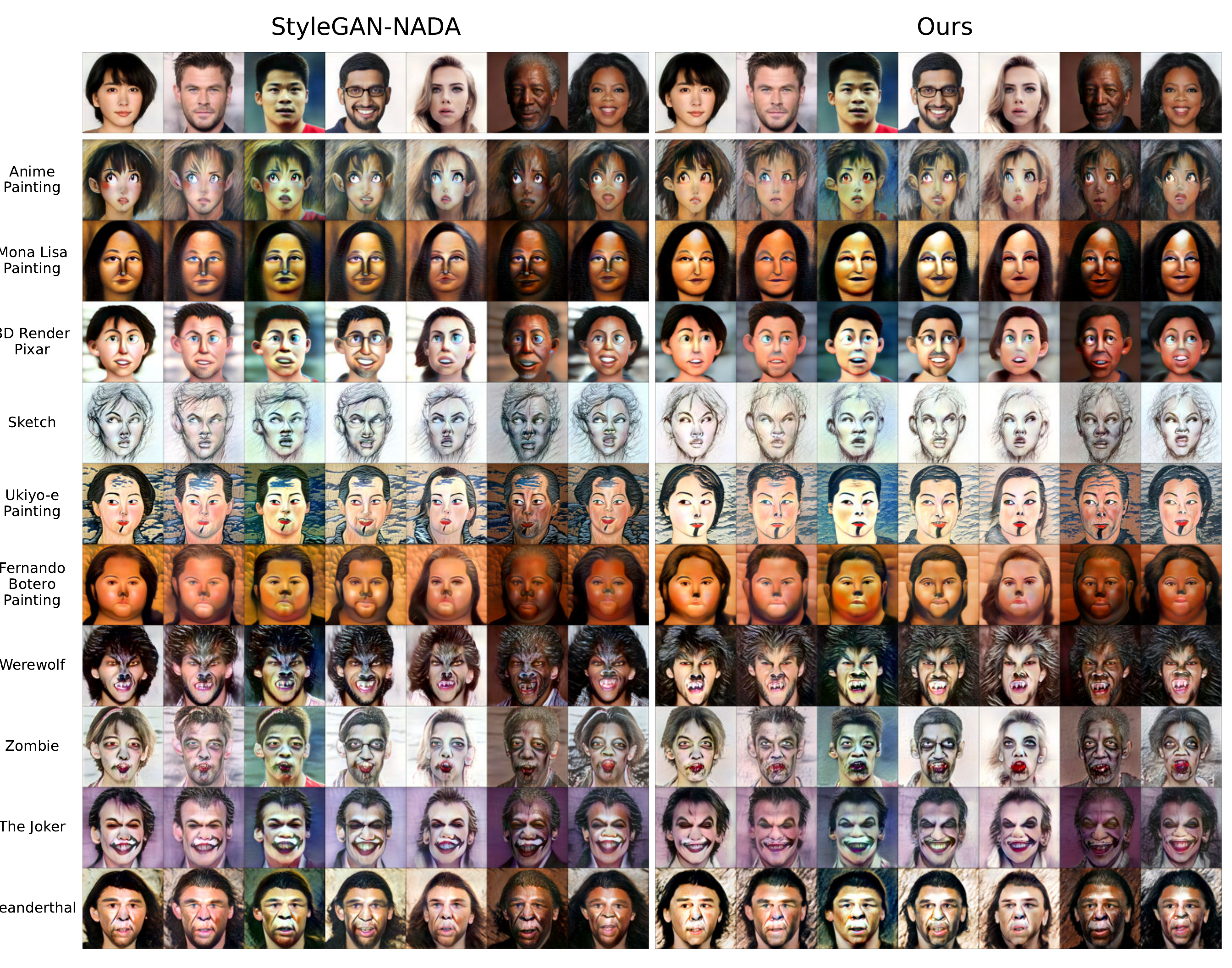}
  \caption{Comparison of text-based domain adaptation methods without indomain angle loss. Left column represents StyleGAN-NADA  \cite{gal2021stylegan}, right column represents our model.}
  \label{fig:text_1}
\end{figure}

\begin{figure}[!h]
  \includegraphics[width=\textwidth]{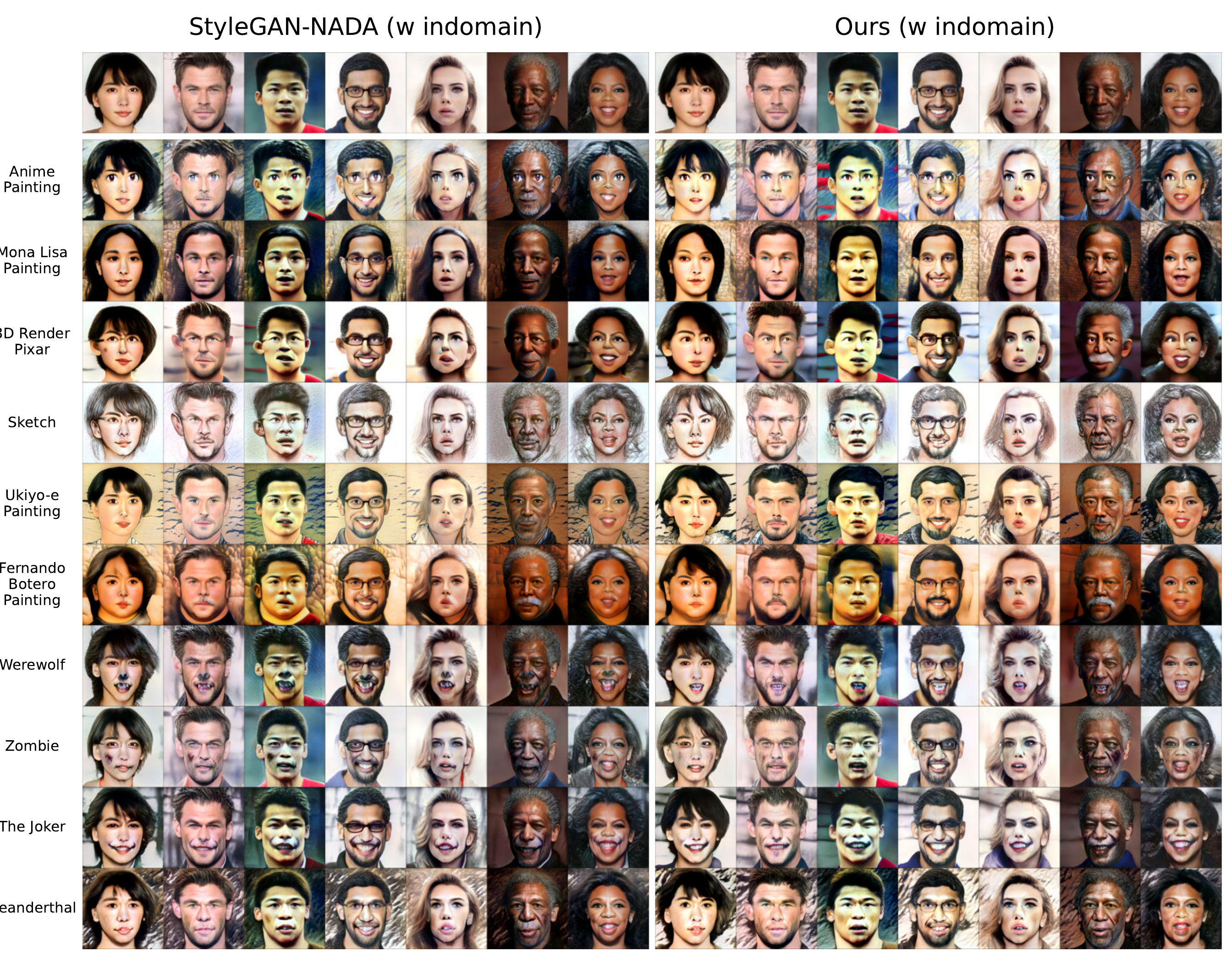}
  \caption{Comparison of text-based domain adaptation methods with indomain angle loss. Left column represents StyleGAN-NADA  \cite{gal2021stylegan}, right column represents our model.}
  \label{fig:text_2}
\end{figure}

Also we report samples for the StyleGAN-NADA and our model with and without indomain angle loss in \Cref{fig:text_1,fig:text_2}. We see that qualitatively indomain angle loss also significantly improves the diversity of the domain adaptation methods. 

\begin{table*}[!b]
\centering
\caption{Evaluation of text-based adaptation methods.}
	\label{table:text-quant}
  \begin{tabular}{ lll }
    \toprule
  Model & Quality & Diversity\\
    \midrule\midrule
    \textbf{Anime Painting} \\
    \midrule
    StyleGAN-NADA \cite{gal2021stylegan} & 0.289 & 0.244 \\
    Ours & 0.284 & 0.305 \\
    StyleGAN-NADA+indomain & 0.256 & 0.401 \\
    Ours+indomain & 0.251 & 0.404 \\
    Multi-Domain+domain\_norm+tt\_direction & 0.260 & 0.256 \\
    \midrule
    \textbf{Zombie} \\
    \midrule
    StyleGAN-NADA \cite{gal2021stylegan} & 0.257 & 0.153 \\
    Ours & 0.264 & 0.275 \\
    StyleGAN-NADA+indomain & 0.261 & 0.354 \\
    Ours+indomain & 0.247 & 0.372 \\
    Multi-Domain+domain\_norm+tt\_direction & 0.258 & 0.191 \\
    \midrule
    \textbf{Across ten domains} \\
    \midrule
    StyleGAN-NADA \cite{gal2021stylegan} & 0.270 $\pm$ 0.032 & 0.196 $\pm$ 0.034 \\
    Ours & 0.256 $\pm$ 0.019 & 0.306 $\pm$ 0.030 \\
    StyleGAN-NADA+indomain & 0.249 $\pm$ 0.018 & 0.394 $\pm$ 0.026 \\
    Ours+indomain & 0.240 $\pm$ 0.018 & 0.398 $\pm$ 0.026 \\
    Multi-Domain+domain\_norm+tt\_direction & 0.247 $\pm$ 0.026 & 0.250 $\pm$ 0.041 \\
    \midrule
    \bottomrule
  \end{tabular}
  \vspace{-0.5cm}
\end{table*}

\subsubsection{Additional Samples}
\begin{figure}[!h]
  \includegraphics[width=\textwidth]{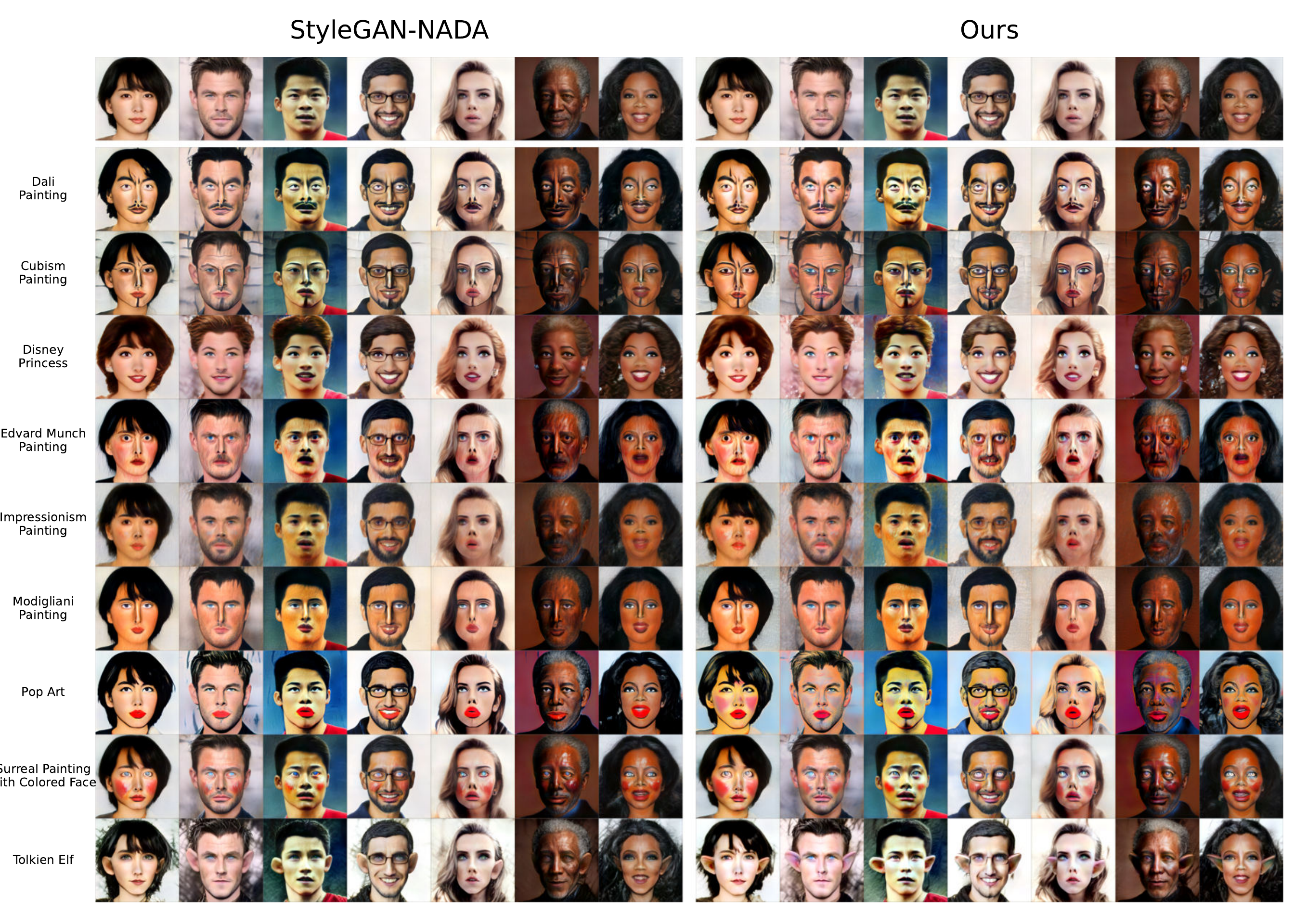}
  \caption{Comparison for other domains in single domain adaptation setup on real images. Left column represents StyleGAN-NADA \cite{gal2021stylegan}, right column represents results with same approach patched with ours parameterization.}
  \label{fig:single_supplementary}
\end{figure}

\begin{figure}[!h]
  \includegraphics[width=\textwidth]{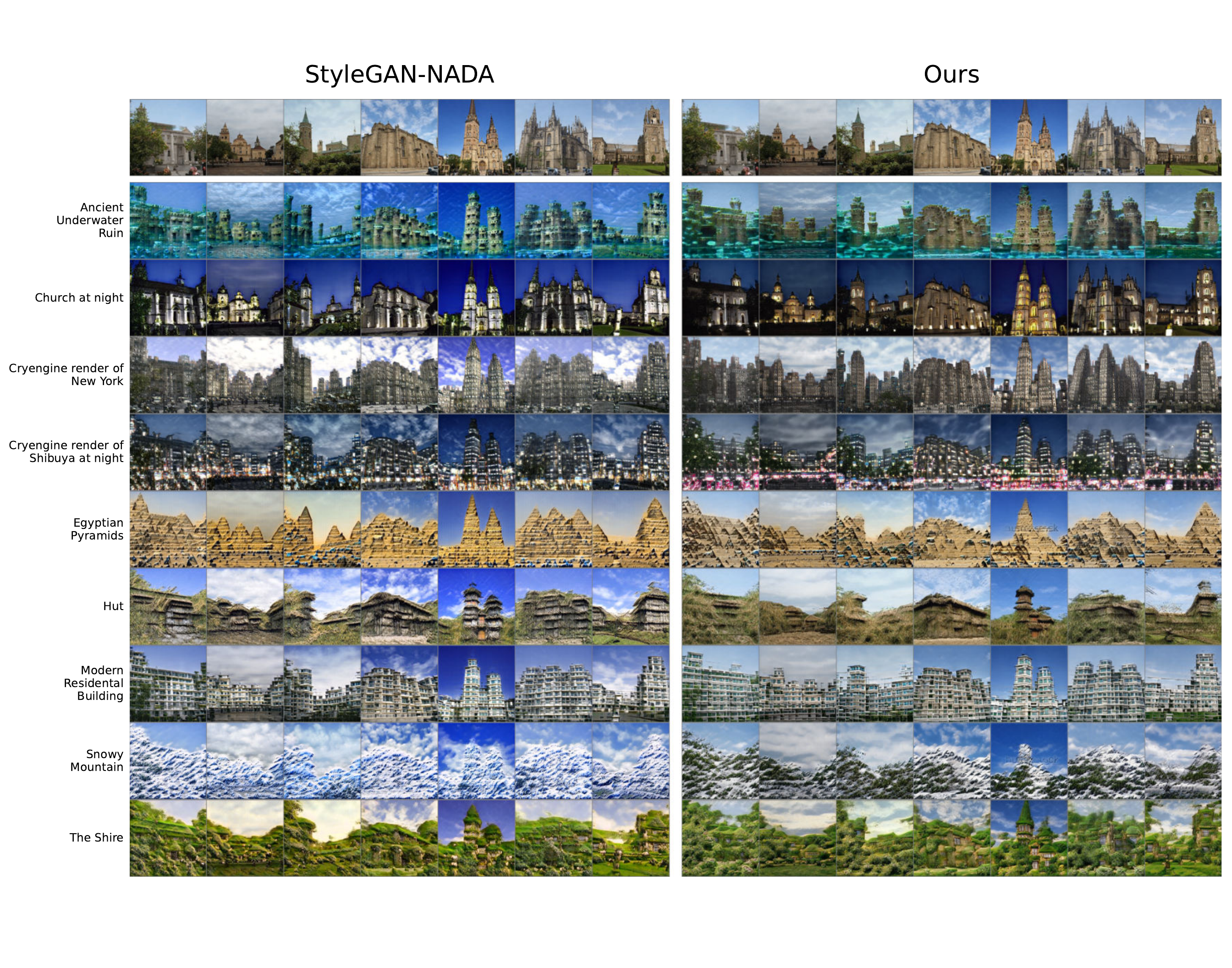}
  \caption{Single domain adaptation comparison for LSUN Church dataset.}
  \label{fig:supplementary_churches_comparison}
\end{figure}

\begin{figure}[!h]
  \includegraphics[width=\textwidth]{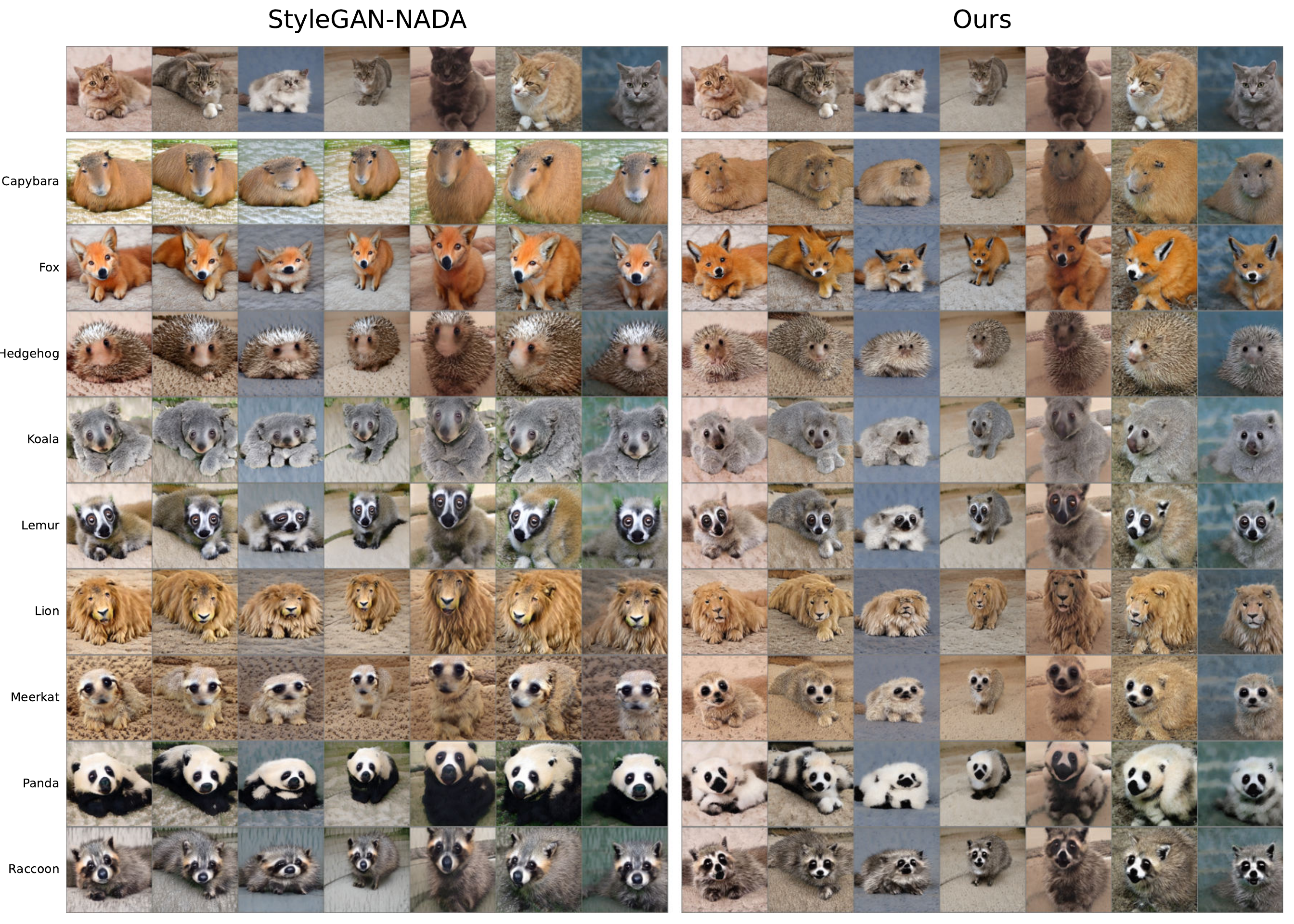}
  \caption{Single domain adaptation comparison for LSUN Cats dataset.}
  \label{fig:supplementary_cats_comparison}
\end{figure}

\begin{figure}[!h]
  \includegraphics[width=\textwidth]{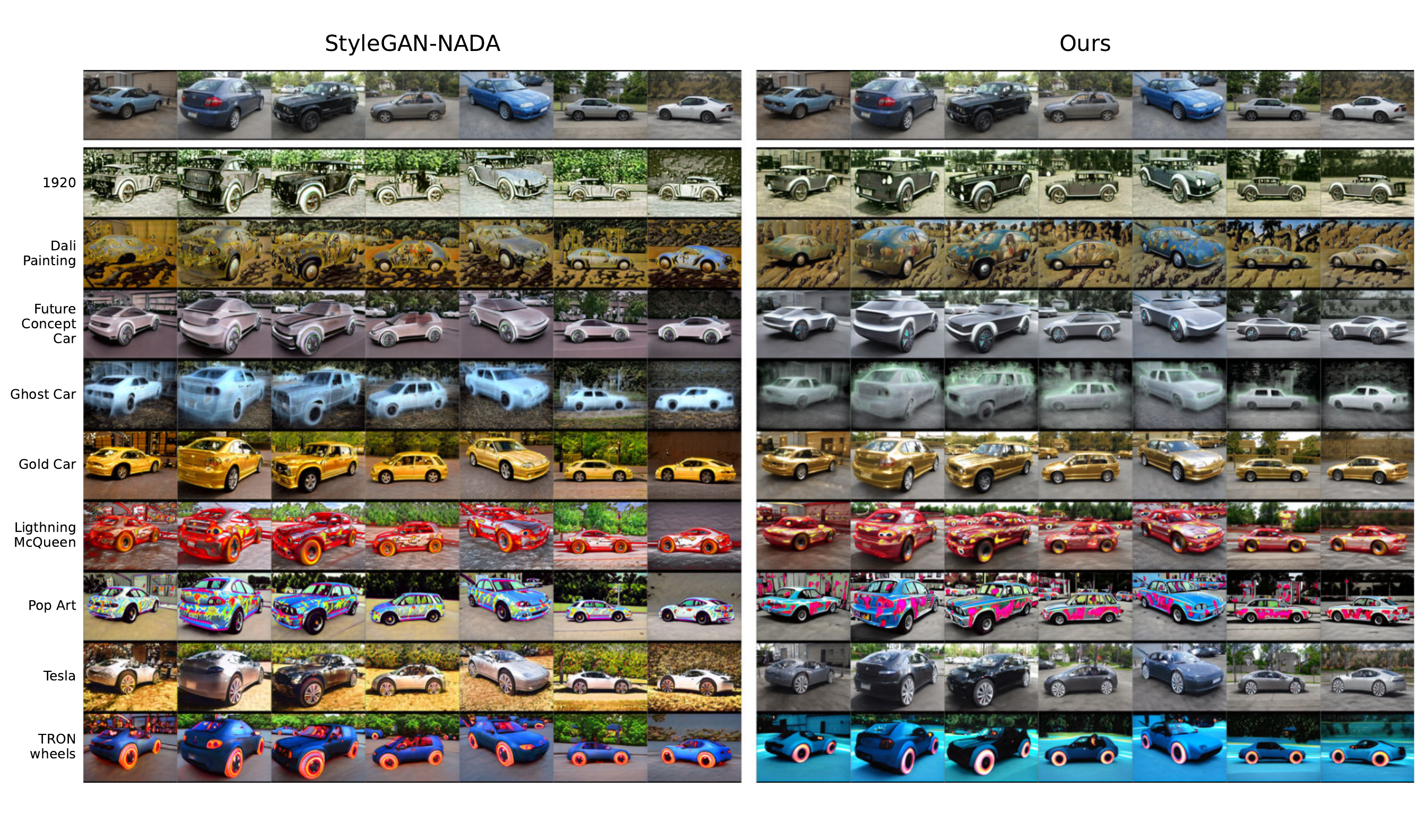}
  \caption{Single domain adaptation comparison for LSUN Cars dataset.}
  \label{fig:supplementary_cars_comparison}
\end{figure}

We show additional domains for FFHQ dataset in \Cref{fig:single_supplementary}. Also we demonstrate how our method works on another datasets such as LSUN Church in \Cref{fig:supplementary_churches_comparison}, LSUN Cats in \Cref{fig:supplementary_cats_comparison}, and LSUN Cars in \Cref{fig:supplementary_cars_comparison}. 

\FloatBarrier

\subsection{Results for One-Shot Domain Adaptation}\label{appx:image_based}
\subsubsection{Hyperparameters}
For each target style image we adapt the generator for 600 iterations as in \cite{zhu2021mind}. We use batch size of 4, fine-tune all layers of the StyleGAN2, set the mixing probability to 0.9. We use all loss terms as in \cite{zhu2021mind} with the same weights and add the $\mathcal{L}_{indomain-angle}$ term with weight 2. For all experiments, we use an ADAM Optimizer with a learning rate of 0.002.

\subsubsection{Training and Inference Time}
The training of the one target style image for 600 iterations on a single Tesla A100 GPU takes about 20 minutes. The same as for the text-based adaptation the inference time consists of two parts: embedding process and the forward pass through the generator. The embedding process takes 0.36 seconds for e4e \cite{tov2021designing} and two minutes for II2S \cite{zhu2020improved}. The second part is the forward pass through adapted GAN generator which works in 0.02 seconds. 

\subsubsection{Additional Samples}
\begin{figure}[!h]
  \includegraphics[width=\textwidth]{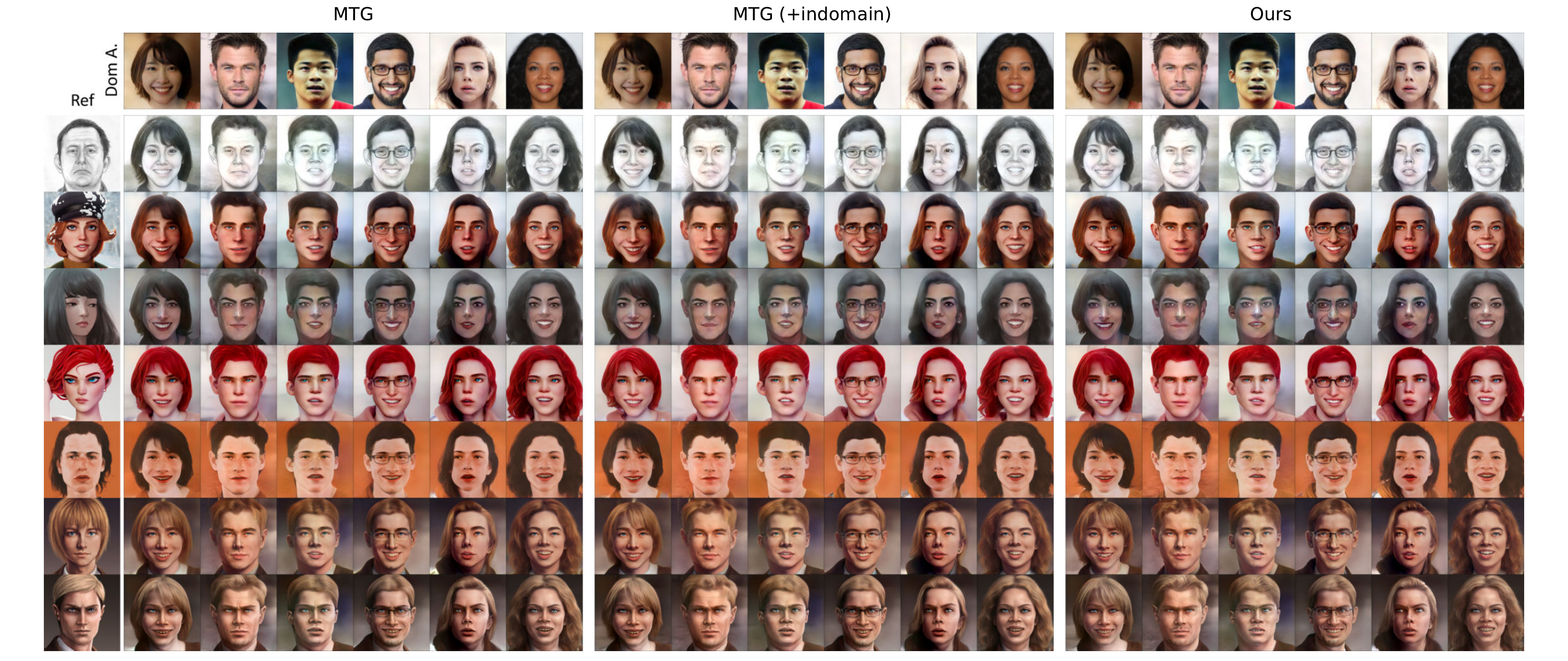}
  \caption{Comparison of one-shot domain adaptation methods: original MindTheGap \cite{zhu2021mind} (left), MindTheGap + indomain (center) and MindTheGap with our parameterization (right).}
  \label{fig:orig_mtg_comparison}
\end{figure}

\begin{figure}[!h]
  \includegraphics[width=\textwidth]{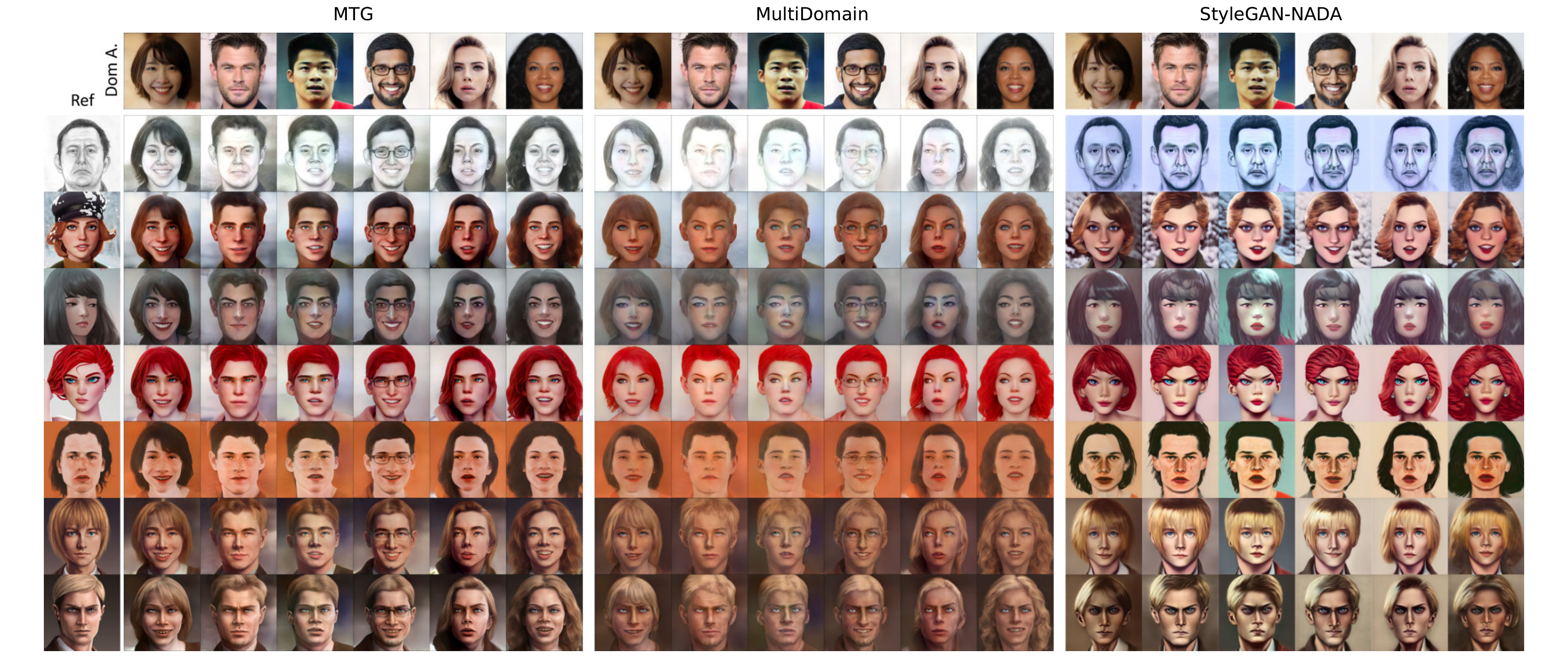}
  \caption{Comparison of one-shot domain adaptation methods: original MindTheGap \cite{zhu2021mind} (left), Multi-Domain model (center) and StyleGAN-NADA (right).}
  \label{fig:mtg_multidomain}
\end{figure}

\begin{figure}[!h]
  \includegraphics[width=\textwidth]{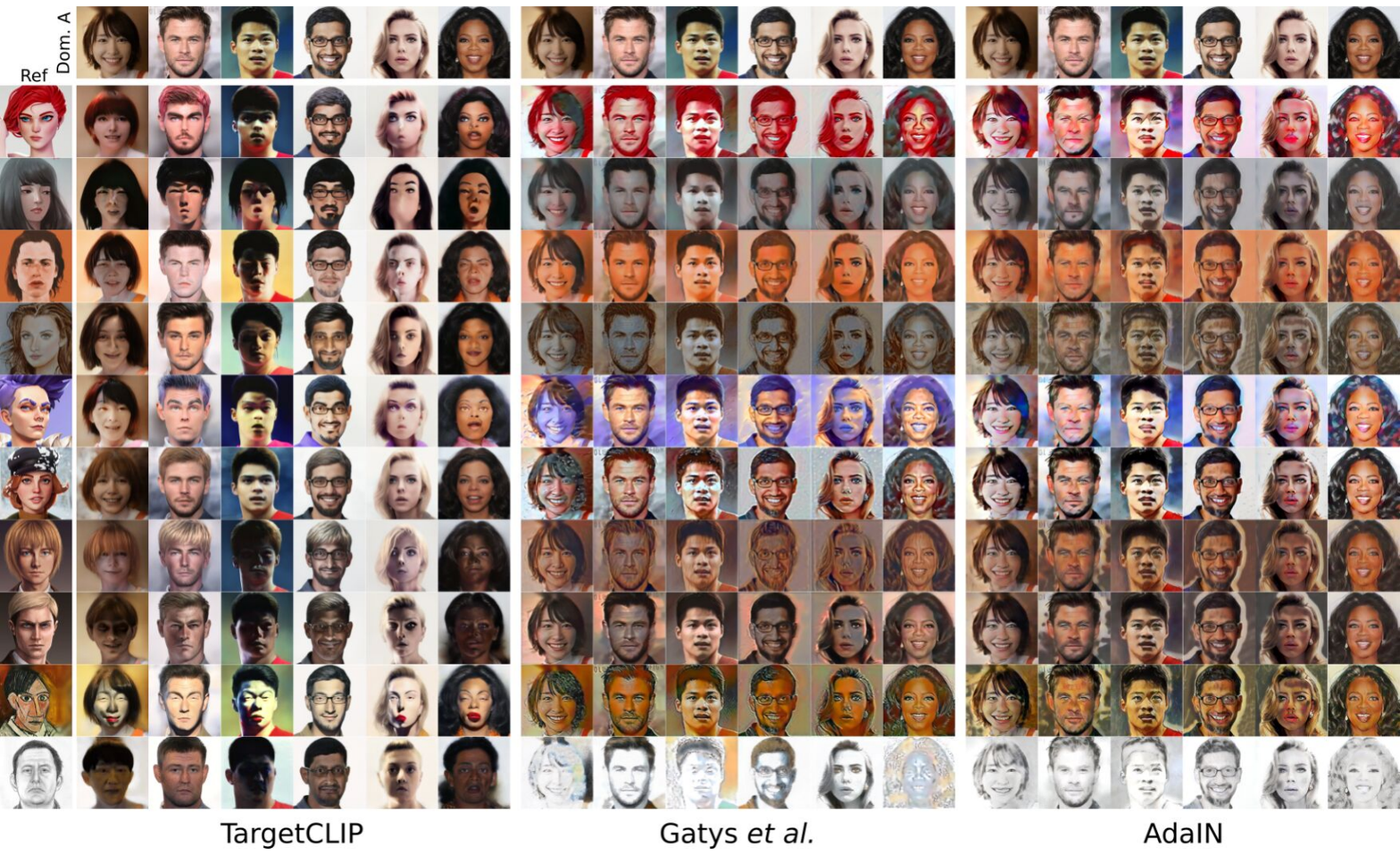}
  \caption{Additional comparisons with other baseline methods including TargetCLIP \cite{chefer2021image}, Gatys et al. \cite{gatys2016image}, and AdaIN \cite{huang2017arbitrary}. Compare these results to our method in \Cref{fig:image_based_comparison}. We can see that both the original MindTheGAP and with our parameterization has fewer artifacts.} 
  \label{fig:other_methods_comparison}
\end{figure}

We provide additional samples in \Cref{fig:orig_mtg_comparison,fig:mtg_multidomain}. 
Also we provide results for other baseline methods in \Cref{fig:other_methods_comparison}.  


\FloatBarrier

\end{document}